\pdfoutput=1

\documentclass[11pt]{article}

\usepackage[preprint]{acl}

\usepackage{times}
\usepackage{latexsym}
\usepackage{fancybox}
\usepackage{booktabs} 
\usepackage{colortbl} 
\usepackage{xcolor}   
\usepackage[most]{tcolorbox}
\usepackage[T1]{fontenc}

\usepackage[utf8]{inputenc}

\usepackage{microtype}

\usepackage{inconsolata}

\usepackage{graphicx}
\usepackage{subcaption}
\usepackage{multirow}
\usepackage{hyperref}

\title{Influences on LLM Calibration: A Study of Response Agreement, Loss Functions, and Prompt Styles}

\author{
 \textbf{Yuxi Xia\textsuperscript{1,2,*}},
 \textbf{Pedro Henrique Luz de Araujo\textsuperscript{1,2}},
 \textbf{Klim Zaporojets\textsuperscript{3}},
 \\
 \textbf{Benjamin Roth\textsuperscript{1,4}}
\\ \\
 \textsuperscript{1}Faculty of Computer Science, University of Vienna, Vienna, Austria, 
 \\
 \textsuperscript{2}UniVie Doctoral School Computer Science, Vienna, Austria
\\
 \textsuperscript{3}Department of Computer Science, Aarhus University, Aarhus, Denmark
\\
\textsuperscript{4}Faculty of Philological and Cultural Studies, University of Vienna, Vienna, Austria
\\
\textsuperscript{*}yuxi.xia@univie.ac.at
\\
\\
}
\begin{document}
\maketitle
\begin{abstract}
Calibration, the alignment between model confidence and prediction accuracy, is critical for the reliable deployment of large language models (LLMs). Existing works neglect to measure the generalization of their methods to other prompt styles and different sizes of LLMs. To address this, we define a controlled experimental setting covering 12 LLMs and four prompt styles. We additionally investigate if incorporating the response agreement of multiple LLMs and an appropriate loss function can improve calibration performance. Concretely, we build Calib-n, a novel framework that trains an auxiliary model for confidence estimation that aggregates responses from multiple LLMs to capture inter-model agreement. To optimize calibration, we integrate focal and AUC surrogate losses alongside binary cross-entropy. Experiments across four datasets demonstrate that both response agreement and focal loss improve calibration from baselines. We find that few-shot prompts are the most effective for auxiliary model-based methods, and auxiliary models demonstrate robust calibration performance across accuracy variations, outperforming LLMs' internal probabilities and verbalized confidences. These insights deepen the understanding of influence factors in LLM calibration, supporting their reliable deployment in diverse applications.
\footnote{Code and data will be released upon acceptance}

\end{abstract}

\begin{figure*}
    \centering
    \includegraphics[width=1.\linewidth]{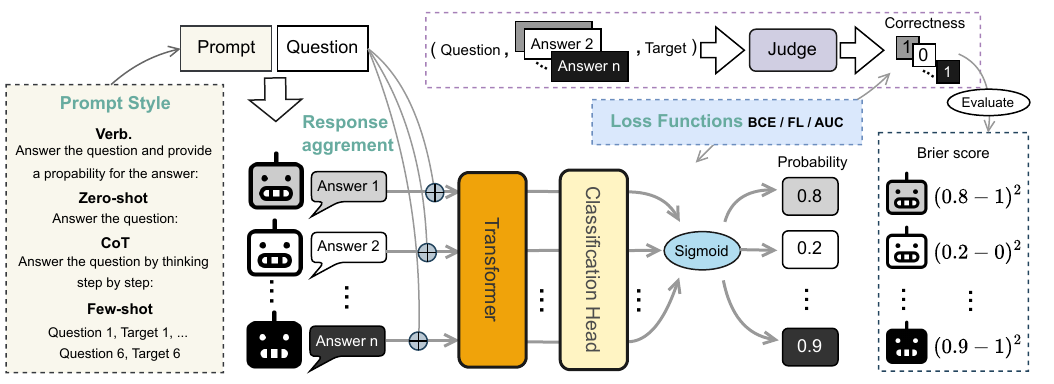}
    \caption{Overview of calibration training of Calib-n  ($n$ indicates the number of LLMs that provide responses). We consider the effect of different \textbf{prompt styles} on calibration and design four diverse prompts to query $n$ target LLMs to provide answers to a question. $n$ joint strings of the question with each answer presenting the \textbf{response agreement} of LLMs are passed to the auxiliary model to generate probabilities for each answer. The auxiliary model is optimized with three \textbf{loss functions} respectively on the correctness of the LLM answers. Brier score (one of four metrics) is used to evaluate the calibration performance of the auxiliary model.}
    \label{fig:calib-n}
\end{figure*}

\section{Introduction}

 Improving the calibration of Large Language Models (LLMs), i.e., aligning the model's confidence with the accuracy of its predictions, can maintain their reliability, usability, and ethical deployment in domains like medicine, law, and education \cite{calib-guo, jiang2021can, geng2024survey}. As LLMs are increasingly integrated into decision-making processes, poor calibration can amplify risks of misinformation, propagate biases, and foster over-reliance among users \cite{raji2020closing}. 

Recent studies \cite{ni2024largelanguagemodelshonest} indicate that LLMs struggle to accurately express their internal confidence in natural language, particularly in the form of verbalized confidence \cite{tian2023just}. \citet{liu2024litcablightweightlanguagemodel} addresses this issue by training a linear layer to adjust the hidden states of the LLM’s final layer for confidence prediction, but this approach only applies to LLMs with accessible weights. In contrast, \citet{ulmer-etal-2024-calibrating} introduces an auxiliary model for confidence estimation using only the generations of the target LLM but is constrained by its evaluation on just two LLMs and two prompt styles. These narrow experimental setups limit the generalization of their findings.

Our study tackles this limitation by comprehensively examining the generalization of different methods over 12 LLMs and four prompt styles. 
Additionally, we investigate the influence of diverse factors on the calibration of LLMs, which includes response agreements among LLMs, three loss functions, and four prompt styles. Concretely, we introduce \textbf{Calib-n} (illustrated in Fig.~\ref{fig:calib-n}), a novel framework aggregating responses from multiple LLMs ($n$ indicates the number of LLMs) to train a single auxiliary model for confidence estimation. Calib-n captures inter-model agreement, mitigating overfitting and reducing overconfidence associated with individual LLMs for calibration \cite{kim-etal-2023-bag}. Except for standard binary cross-entropy (BCE) loss, we incorporate focal (FL) \cite{8417976} and AUC surrogate (AUC) \cite{9710509} loss functions that are effective for improving the calibration of non-transformer type of neural networks \cite{mukhoti2020calibrating, pmlr-v119-moon20a}. 
Finally, we systematically study the effects of prompt styles, testing four diverse prompt types: Verbalized \cite{tian2023just}, Chain-of-Thought (CoT) \cite{wei2022chain}, Zero-shot, and Few-shot prompts.


Our experiments span four open-ended question answering datasets and 12 LLMs, including seven small models (2--9B parameters) and five large models (27--72B parameters) from five distinct model families. The results indicate that no single method consistently outperforms all others across the various models, datasets, and prompt configurations. Yet, after aggregating the results and counting the number of overall wins for each method, we uncover new insights regarding the calibration factors of the analyzed settings: 
\begin{itemize}

    \item \textbf{Response Agreement:} By leveraging inter-model response agreement, Calib-n outperforms the state-of-the-art baselines.

    \item \textbf{Loss Functions:} FL loss improves calibration compared to BCE and AUC losses, demonstrating its effectiveness for both Calib-1 (i.e., using responses from one LLM) and Calib-n. Calib-1 trained with FL yields the best results. 

    \item \textbf{Prompt Style:} We find that the effectiveness of methods is highly influenced by the prompt styles, with few-shot prompts proving to be the most beneficial for training auxiliary models.

    \item \textbf{Accuracy-Calibration Correlation:} With accuracy variances caused by different dataset complexities, prompt styles and models, we find that auxiliary models maintain robust calibration performance across accuracy changes, in contrast to the fluctuating calibration of LLMs that rely on internal probabilities and verbalized confidences.
\end{itemize}
Our findings underscore the importance of reexamining calibration strategies for LLMs, suggesting the use of response agreement, focal loss and few-shot prompts to improve the calibration of LLMs. 

\section{Related Work}

\noindent \textbf{Calibration} 
The concept of calibration of neural networks was introduced by \citet{guo2017calibration}. 
\citet{lin2022teaching} show that GPT-3 can learn to express uncertainty about its own answers without the use of logits. Later, \citet{tian2023just} demonstrate that verbalized confidence is generally better calibrated than the conditional probabilities w.r.t the consistency of the LLMs. However, \citet{zhang-etal-2024-dont-go} show that LLM probabilities and verbalized confidence tend to overly concentrate within a fixed range. Similarly, \citet{ni2024largelanguagemodelshonest} analyze and compare probabilistic and verbalized perceptions of the knowledge boundaries of LLM, highlighting their challenges in confidence estimation. 
To address these issues, \citet{liu2024litcablightweightlanguagemodel} train a linear layer to adjust the last layer’s hidden states of LLMs for confidence generation.  \citet{ulmer-etal-2024-calibrating} propose a method to estimate LLM confidence based only on textual input and output.  
However, none of these works perform a comprehensive analysis of \textbf{different influence factors in LLM calibration}. These factors can be loss functions, response agreement and prompt styles. \citet{mukhoti2020calibrating} prove that focal loss \cite{8417976} can improve the calibration of neural networks. AUC surrogate loss \cite{9710509} and correctness ranking loss \cite{pmlr-v119-moon20a} calibrate models by adjusting the ranking of logits to ensure positive samples are ranked higher than negative samples.
\citet{kim-etal-2023-bag} suggest combining the predictions from multiple models (response agreement) can mitigate overfitting and reduce overconfidence for calibration.
\citet{min2022rethinking, wang2023label, chen2024unleashingpotentialpromptengineering} showcase that LLMs output is highly dependent on the prompt and different prompt styles can significantly influence the performance of LLMs. 
Our work comprehensively studies the influence of response agreement, three loss functions, and four prompt styles with 12 LLMs on four open-ended question answering datasets.


\section{Methodology}
Fig.~\ref{fig:calib-n} demonstrates our framework, which includes four prompt styles, assessment of the correctness of LLM generation, and calibration training of auxiliary models for confidence estimation and calibration evaluation for these models.

\subsection{Prompt Styles}
Prompt styles significantly impact LLMs' performance \cite{chen2024unleashingpotentialpromptengineering}. \citet{liu2024litcablightweightlanguagemodel} use Few-shot prompts to query LLM answers for calibration experiments. \citet{tian2023just} employ verbalized prompts to elicit probability estimates from LLMs regarding their responses. \citet{ulmer-etal-2024-calibrating} evaluate their methods on both CoT and verbalized prompts. However, these studies either evaluate their methods against baselines using inconsistent prompt styles or fail to provide a comprehensive comparison across diverse prompt styles. For example, \citet{liu2024litcablightweightlanguagemodel} use verbalized prompts to generate verbalized probabilities as a baseline, but apply few-shot prompts to their method, which ignores the potential influence of prompts in calibration evaluation. Our work comprehensively studies the impact of prompt styles in LLM calibration and employs the most commonly used prompts: Verbalized (Verb.), Zero-shot, CoT, and Few-shot prompts. The detailed prompts are shown in Appendix ~\ref{app:llm_prompts}. Given a question $q$ with one of the prompts, we process the answers generated by $n$ target LLMs with regular expression-based text processing.




\subsection{Correctness of LLM Generation }
We employ a Judge model $\mathcal{J}$, Prometheus-8x7b-v2.0, to assess the correctness of generated answers w.r.t target answers.  We selected this model because it is open-source and its judgments have been shown to strongly correlate with those of human evaluators and large proprietary models~\cite{kim2024prometheus2opensource}.
Given an input question $q$, a target answer $y$, and a generated answer $a_{i}$ by the $i$-th target LLM $\mathcal{M}_i$, $\mathcal{J}$ is prompted to provide a binary correctness score $c_{i}$ that reflects the semantical equivalence between $a_{i}$ and $y$. The specific prompt is shown in Appendix \ref{app:judge_prompts}.
\begin{equation}
    c_{i} = \mathcal{J}(a_{i} \overset{\text{semantic}}{=} y|q, y, a_{i}), ~ c_{i}\in\{0,1\}
\end{equation}


\subsection{Response Agreements of Multiple LLMs}

Different from previous work \cite{liu2024litcablightweightlanguagemodel, ulmer-etal-2024-calibrating, tian2023just} that estimate confidence using the information from a single target LLM, Calib-n leverages the responses from $n$ target LLMs to train an auxiliary model for estimating jointly the confidence for each LLM. This setting is inspired by \citet{kim-etal-2023-bag} which suggests that combining predictions can mitigate overfitting and reduce overconfidence inherent to individual models for confidence estimation.
By having access to the responses of $n$ LLMs, the auxiliary model can infer cases of low consensus among models, signaling increased uncertainty. We verify the effectiveness of this setting by comprehensively comparing the results of using the generations produced by one LLM (\textbf{Calib-1}) to using the generations from multiple LLMs (\textbf{Calib-n}). 

 The auxiliary model, $f(\cdot)$, is composed of a transformer backbone (bert-base-uncased \cite{DBLP:journals/corr/abs-1810-04805}), a classification head ($768\rightarrow n$) and a sigmoid activation function, which outputs probabilities $P$ for LLM answers $\{a_i\}_{i=1}^n$. The input for $f(\cdot)$ is $n$ joint strings of $q$ with each answer (e.g., $q$[SEP]$a_i$), denoted as $\{q+a_i\}_{i=1}^n$. 
\begin{equation}
    P = f(\{q+a_i\}_{i=1}^n) =\{p_i\}_{i=1}^n, ~p_i\in[0,1]
\end{equation}
 
\textbf{Training objective}. The goal is to optimize the predicted probability to align with the correctness of the input answer. Give $k$ questions, we minimize the averaged BCE loss of each LLM answer:
\begin{equation}
\begin{split}
\mathcal{L}_{BCE} = - \frac{1}{k \cdot n } 
& \sum_{j=1}^{k}  \sum_{i=1}^{n} \Big( c^{(j)}_i \log(p^{(j)}_i) +  \\
&(1 - c^{(j)}_i) \log(1 - p^{(j)}_i) \Big)
\end{split}
\end{equation}


\textbf{Evaluation}. A target LLM $\mathcal{M}_i$ achieves a low calibration Brier score if $p_i$ accurately reflects the reliability of $a_i$, which is the correctness $c_i$. Specifically, for all $k$ generated answers of $\mathcal{M}_i$ regarding $k$ questions, the Brier score~\cite{brier1950verification} of  $\mathcal{M}_i$ average squared error between all predicted probabilities and correctness of these answers:
\begin{equation}
\label{eq:brier}
    Brier(\mathcal{M}_i) = \frac{1}{k}\mathop{\sum}^k_{j=1}(p^{(j)}_i -c^{(j)}_i)^2 
\end{equation}
Following \citet{tian2023just}, we also evaluate all methods with three other metrics (details in Section \ref{eval_metrics}).

\subsection{Loss Functions}
To further improve the calibration, we experiment with focal and AUC losses in addition to BCE loss. 

\textbf{Focal loss (FL)} \cite{8417976, mukhoti2020calibrating} focuses on hard-to-classify examples, reducing the weight of correctly classified samples and encouraging the model to focus on predictions with high BCE loss. This loss is commonly used in imbalanced datasets but can benefit calibration since it emphasizes predictions with a large discrepancy between confidence and correctness. The equation is:
\begin{equation}
\mathcal{L}_{\text{FL}} = - \frac{1}{k} \sum_{j=1}^{k} \Big(\alpha (1 - e^{-\mathcal{L}_{BCE_{j}}})^\gamma \cdot \mathcal{L}_{BCE_{j}} \Big)
\end{equation}
Where $\mathcal{L}_{BCE_{j}}$ is the BCE loss of the data sample $j$. We use the default setting from \citet{8417976} to set $\alpha = 0.25$ and $\gamma=2.0$.

\textbf{AUC Surrogate loss (AUC)} \cite{9710509} uses a logistic loss to maximize the differences between true and false answers' scores (logits $x$ generated by the classification head). The equation is:
\begin{equation}
    \mathcal{L}_{\text{AUC}} = \frac{1}{|T| \cdot |F|} \sum_{t \in T} \sum_{f \in F} \sigma\left( x_f - x_t \right)
\end{equation}
Where $T$ and $F$ are the indexes set for all true and false answers respectively. $\sigma$ stands for sigmoid function.

In the end, we propose the following methods combining different techniques and verify their effectiveness with comprehensive experiments across different models, datasets and prompts.

\textbf{(BCE)/(FL)/(AUC)Calib-1}: calibration training using the generations from one target LLM and optimizing with BCE/FL/AUC loss function.

\textbf{(BCE)/(FL)/(AUC)Calib-n}: calibration training using the generations from $n$ target LLM and optimizing with BCE/FL/AUC loss function.

\textbf{(BCE)/(FL)/(AUC)Calib-n+PS}: Platt Scaling (PS) ~\cite{platt1999probabilistic} (explained in Section \ref{baselines}) rescales the probabilities of test data by learning on the probabilities of validation data. Those probabilities are generated by corresponding Calib-n models.


\section{Experiments}
To comprehensively compare confidence estimation methods, we include diverse datasets, LLMs, and state-of-the-art baselines in our experimental setting.

\subsection{Datasets}
We cover four open-ended quenstion-answering datasets: TriviaQA~\cite{joshi2017triviaqa}, Sciq~\cite{SciQ}, WikiQA~\cite{yangwikiqa2015}, and NQ~\cite{kwiatkowski2019natural}.
We adopt the setting from \citet{liu2024litcablightweightlanguagemodel}, where 2k/1k samples are selected as training/test data for TriviaQA, Sciq, and NQ, and 1040/293 for WikiQA.

\subsection{Models}
We include 12 models from five families: Llama~\cite{grattafiori2024llama3herdmodels,touvron2023llama2openfoundation}, Phi~\cite{abdin2024phi3technicalreporthighly}, Gemma~\cite{gemmateam2024gemma2improvingopen}, Qwen~\cite{yang2024qwen2technicalreport}, and Mixtral~\cite{jiang2024mixtralexperts}.\footnote{All models are available at \url{https://huggingface.co/models}.}
In our analyses, we cluster the models based on their number of parameters: 

\textbf{Small models (2-9B parameters):} 
Llama-2-7b-chat-hf (referred as Llama2-7b), Llama-3.1-8B-Instruct (Llama3.1-8b), Llama-3-8B-Instruct (Llama3-8b), Phi-3-small-128k-instruct (Phi3-7b), Phi-3.5-mini-instruct (Phi3-4b), gemma-2-2b-it (Gemma2-2b), gemma-2-9b-it (Gemma2-9b).

\textbf{Large models (27-72B parameters):}  Qwen2-72B-Instruct (Qwen2-72b), Llama-3-70B-Instruct (Llama3-72b), Llama-3.1-70B-Instruct (Llama3.1-70b), Mixtral-8x7B-Instruct-v0.1 (Mixtral-8x7b), gemma-2-27b-it (Gemma2-27b).

Calib-n models are trained with all LLMs inside the same group and provide confidence scores for each model in that group. 






\subsection{Evaluation Metrics} \label{eval_metrics}


We report four metrics for calibration evaluation:

\textbf{ECE:} The expected calibration error~\cite{guo2017calibration} is computed by partitioning the predictions into 10 bins based on their confidence and then taking the weighted (by the number of samples in a bin) average of the squared difference between bin average accuracy and confidence.

\textbf{ECE-t:} The temperature-scaled expected calibration error~\cite{tian2023just} finds a single temperature scaling parameter $\beta$ that minimizes the negative log-likelihood between model confidences and answer correctness.
Then, $\beta$ is used to scale the confidences before the ECE is computed.

\textbf{Brier:} The Brier Score~\cite{brier1950verification} is the average squared error between predictions' confidence and correctness (see Eq.~\ref{eq:brier}).

\textbf{AUC:} The area under the curve (AUC) of selective accuracy and coverage~\cite{geifman2017selective}.
Let coverage be the fraction of all test samples with a confidence lower than some threshold $t, 0 \leq t \leq 1$. The accuracy-coverage curve plots the coverage against the accuracy (of covered samples) for all threshold values. 

\paragraph{Aggregate analysis.}
We aggregate the performance of different confidence estimation methods by counting their number of \textbf{wins}: given each metric above, we count the number of times a method outperforms the others in all possible combinations of prompt style, dataset, and model.

\subsection{Baseline Methods} \label{baselines}
We compare Calib-* with standard baselines and the state-of-the-art confidence estimation methods:

\textbf{LLM Probabilities (LLM Prob.):} The conditional sequence probability $P_\theta\left(y|x\right)$ of an answer $y$ given an input $x$, according to the model parametrized by $\theta$.

\textbf{LLM Prob. + Platt scaling (PS):} This method applies Platt scaling~\cite{platt1999probabilistic} to the previous baseline.
That is, two scalars $a,b \in \mathcal{R}$ are used to scale the original LLM probability $p$: $p_{\text{ps}} = \sigma\left(ap + b\right)$, where $\sigma$ is the sigmoid function.
We obtain parameters $a$ and $b$ by minimizing the mean-squared error between model confidences and answer correctness.

\textbf{Verbalized confidences (Verbalized \%) \cite{tian2023just}:} The probability of correctness expressed in models' (text) responses given the Verb. prompts.

\textbf{APRICOT \cite{ulmer-etal-2024-calibrating}:} A recent method for calibrating LLMs. It consists of clustering related questions and measuring the per-cluster accuracies given answers from a target LLM. Then the cluster accuracies are used as the references to train an auxiliary transformer model that outputs confidence values for the target LLM.

\begin{figure}
    \centering
    \includegraphics[width=1\linewidth]{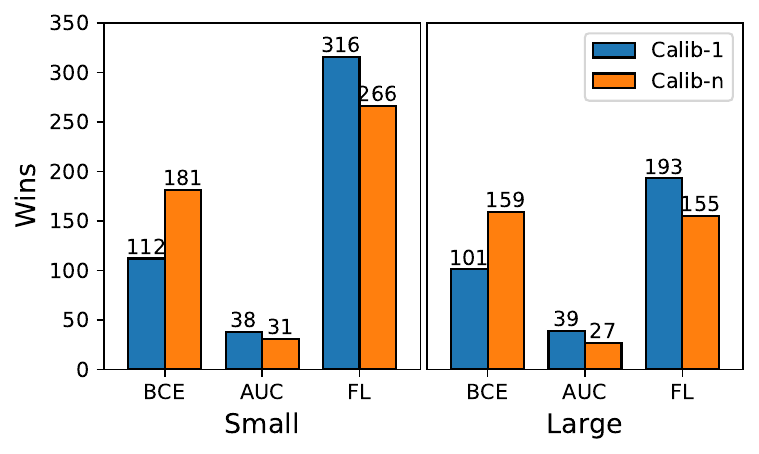}
    \caption{Comparison of Calib-1 and Calib-n methods based on the number of wins across different loss functions for calibrating small (left) and large (right) LLMs.}
    \label{fig:loss_comparison}
\end{figure}

\begin{figure*}[ht]
  \centering
    \begin{subfigure}{0.272\textwidth}
        \centering
        \includegraphics[width=1\linewidth]{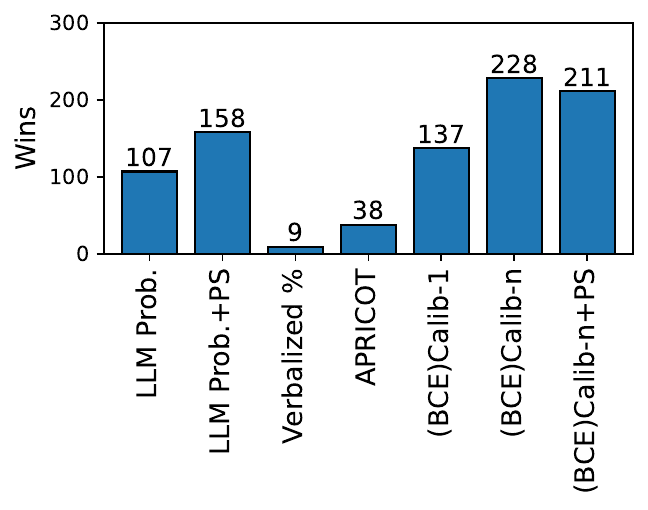}
        \caption{(BCE)Calib-* vs Baselines}\label{fig:main_bce}
    \end{subfigure}
    \begin{subfigure}{0.263\textwidth}
        \centering
        \includegraphics[width=1\linewidth]{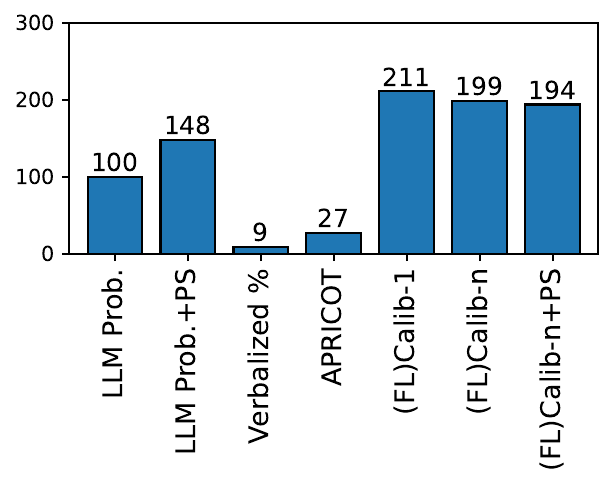}
        \caption{(FL)Calib-* vs Baselines}\label{fig:main_fl}
    \end{subfigure}
    \begin{subfigure}{0.247\textwidth}
        \centering
        \includegraphics[width=1\linewidth]{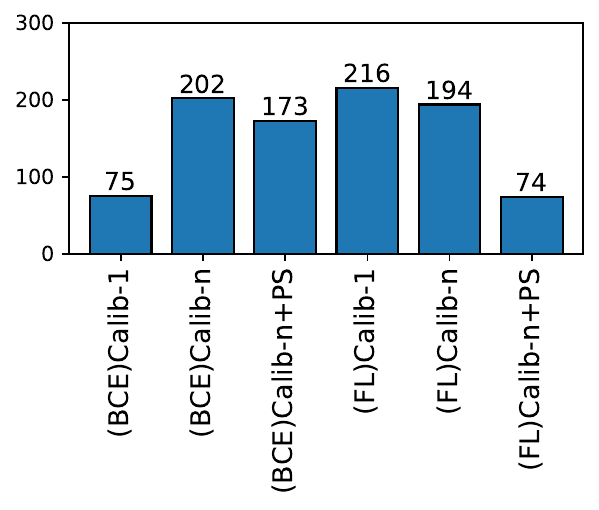}
        \caption{(BCE) vs (FL)Calib-*}\label{fig:main_all}
    \end{subfigure}
    \begin{subfigure}{0.2\textwidth}
        \centering
        \includegraphics[width=1\linewidth]{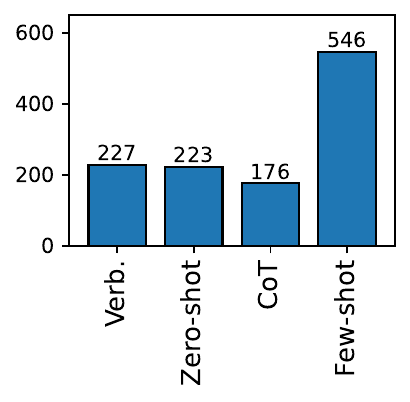}
        \caption{Prompts}\label{fig:main_prompt}
    \end{subfigure}
    \caption{The winning comparison results of different methods and prompts: \ref{fig:main_bce} and \ref{fig:main_fl} sub-figures present the superior results of Calib-* methods using BCE and FL loss respectively when against baselines. \ref{fig:main_all} shows the comparison results among all Calib-* methods, demonstrating that (FL)Calib-1 achieves the best overall performance. \ref{fig:main_prompt} compares the winning result among all prompt styles and shows that using few-shot prompts is the most beneficial.}
    \label{fig:main}
\end{figure*}


\begin{figure}[ht]
    \centering
    \includegraphics[width=1\linewidth]{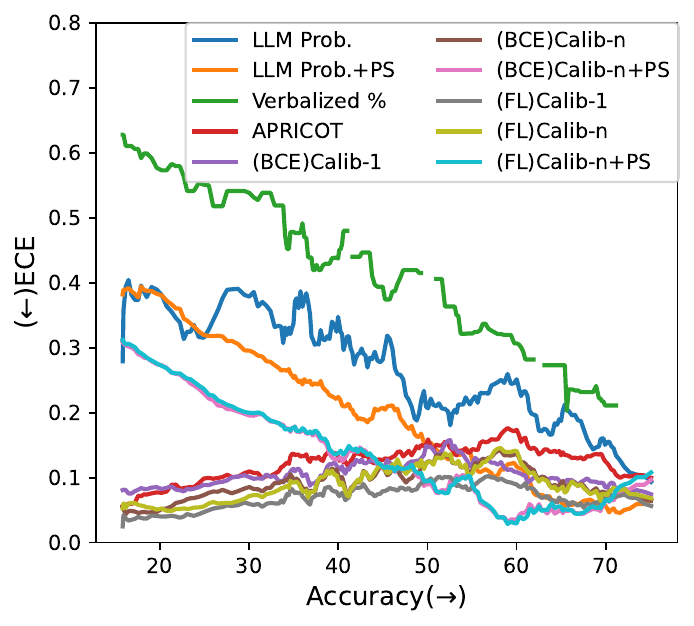}
    \caption{The correlation between accuracies achieved by different configurations (i.e., prompts, models, datasets) and corresponding ECE scores evaluated on different methods. The line of Verbalized \% is not continuous because it only applies to Verb. prompts and thus has fewer accuracy points than other methods. The result indicates that Calib-* and APRICOT are robust to accuracy variations.  Different methods achieve the lowest ECE scores in different accuracy ranges.}\label{fig:acc_ece}
\end{figure}

\begin{table*}[ht!]
\centering
\scriptsize
\renewcommand{\arraystretch}{1.15} 

\begin{tabular}{l@{\hspace{0.1cm}} l@{\hspace{0.15cm}}|l@{\hspace{0.2cm}}l@{\hspace{0.1cm}}l@{\hspace{0.1cm}}l@{\hspace{0.1cm}}|l@{\hspace{0.2cm}}l@{\hspace{0.1cm}}l@{\hspace{0.1cm}}l@{\hspace{0.1cm}}|l@{\hspace{0.2cm}}l@{\hspace{0.1cm}}l@{\hspace{0.1cm}}l@{\hspace{0.1cm}}|l@{\hspace{0.2cm}}l@{\hspace{0.1cm}}l@{\hspace{0.1cm}}l}
\toprule
 & \multirow{3}{*}{\textbf{Method}} & \multicolumn{4}{c}{\textbf{Verb.}} & \multicolumn{4}{c}{\textbf{Zero-shot}} & \multicolumn{4}{c}{\textbf{CoT}} & \multicolumn{4}{c}{\textbf{Few-shot}} \\ 

 \cline{3-18}
 &  & \textbf{\tiny ECE$\downarrow$} & \textbf{\tiny ECE-t$\downarrow$} & \textbf{\tiny Brier $\downarrow$} & \textbf{\tiny AUC$\uparrow$} 
    & \textbf{\tiny ECE $\downarrow$} & \textbf{\tiny ECE-t$\downarrow$} & \textbf{\tiny Brier $\downarrow$} & \textbf{\tiny AUC$\uparrow$} 
    & \textbf{\tiny ECE $\downarrow$} & \textbf{\tiny ECE-t$\downarrow$} & \textbf{\tiny Brier $\downarrow$} & \textbf{\tiny AUC$\uparrow$}  
    & \textbf{\tiny ECE $\downarrow$} & \textbf{\tiny ECE-t$\downarrow$} & \textbf{\tiny Brier $\downarrow$} & \textbf{\tiny AUC$\uparrow$} \\ 
    
\hline
 \multirow{10}{*}{\rotatebox{90}{Gemma2-27b}} & LLM Prob. & \cellcolor[HTML]{FCE3A4}0.445 & \cellcolor[HTML]{FEF6BB}0.249 & \cellcolor[HTML]{FBDA98}0.417 & \cellcolor[HTML]{EC7014}0.704 & \cellcolor[HTML]{FEF6BB}0.447 & \cellcolor[HTML]{FEF6BB}0.255 & \cellcolor[HTML]{FEF6BB}0.419 & \cellcolor[HTML]{EC7014}0.713 & \cellcolor[HTML]{FEF6BB}0.395 & \cellcolor[HTML]{FEF6BB}0.268 & \cellcolor[HTML]{FEF6BB}0.392 & \cellcolor[HTML]{F6B86D}0.671 & \cellcolor[HTML]{FEF6BB}0.235 & \cellcolor[HTML]{FEF6BB}0.184 & \cellcolor[HTML]{FEF6BB}0.308 & \cellcolor[HTML]{FFF7BC}0.584 \\
\hline \hline

\end{tabular}
\caption{Test performance of our methods (Calib-*) and baseline methods on NQ dataset using four different prompts (Verb., Zero-shot, CoT, Few-shot). Calib-n is trained with the responses of all LLMs in the table. Only Verb. prompt asks the LLM to provide a probability for a given answer and thus have the results for Verbalized \% performance. We color the text with a scale normalized by the values gap in each column, with darker shades indicating better performance. The results of the other three datasets and seven models are shown in Appendix \ref{other_calibration}.}\label{tb:nq_large}
\end{table*}

\section{Results and Analysis}
The detailed results (Table \ref{tb:nq_large}-\ref{tb:nq_small}) indicate that no single method consistently outperforms all others across various models, datasets, and prompt configurations. Therefore, we first present the aggregated results, which summarize all the results, followed by a more detailed discussion.

\subsection{Aggregated Result}

\textbf{What is the best loss function for improving calibration?} In Fig. \ref{fig:loss_comparison}, we compare the performance of Calib-1 and Calib-n when optimizing with different loss functions. We observe that \textbf{FL loss wins in most settings}, followed by BCE loss. 
Additionally, we notice that BCE loss outperforms the FL loss when applied in the Calib-n method for large models.

\textbf{What is the best overall method?} Fig. \ref{fig:main_bce} and \ref{fig:main_fl} present the winning comparison results of (BCE)Calib-* and (FL)Calib-* methods against baseline methods respectively. 
The results demonstrate that Calib-* methods gain more wins than baselines in both sub-figures. Verbalized confidences get the lowest number of wins. 
Applying Platt Scaling can further improve the calibration performance of LLM probabilities but this technique is not generalizable to enhance the calibration of Calib-n. Fig. \ref{fig:main_bce} shows that (BCE)Calib-n gains the highest wins against the baseline methods.
While (FL)Calib-1 accrues more wins in Fig. \ref{fig:main_fl}. To identify the best method, we present Fig. \ref{fig:main_all} to compare the performance among our Calib-* methods, demonstrating that \textbf{(FL)Calib-1 exhibits the best overall performance}.

\textbf{Which prompt style is most effective?} The differences in prompt styles are well-known for their impact on the performance of LLMs. Fig. \ref{fig:main_prompt} showcases that prompt styles can also impact calibration performance. The results indicate that \textbf{few-shot is the most beneficial prompt} contributing to the highest wins among all other prompt styles.

\textbf{Which calibration methods maintain robustness to accuracy variations?} Previous work \cite{zhang-etal-2024-dont-go} reveals that confidence estimation methods like LLM probabilities and Verbalized confidence excessively concentrate on a fixed range (tend to be overconfident) and remain unchanged regardless of the dataset’s complexity. To analyze this issue, we test all the confidence estimation methods with different datasets, prompts, and models. Each setting combination (e.g., using Zero-shot prompts to test the TriviaQA dataset on Gemma2-27b model) can result in one single accuracy value and multiple ECE scores---one for each confidence estimation method. We sort the accuracy values from all setting combinations and analyze the correlation between the accuracies and ECE scores. The results are presented in Fig. \ref{fig:acc_ece}. We observe that ECE scores of LLM probabilities, LLM Prob.+PS, Verbalized confidences and (BCE)Calib-n+PS are highly correlated with accuracies, i.e., the ECE scores decrease when the accuracies get higher which verified the findings of \citet{zhang-etal-2024-dont-go}.  \textbf{APRICOT, Calib-1 and Calib-n are robust to the accuracy changes}, the ECE scores of these methods remain relatively stable across different accuracies. Verbalized confidence consistently shows the lowest performance across all accuracy levels.

\textbf{What is the best method for different accuracy levels?} The optimal goal of current state-of-the-art confidence estimation methods should not only focus on achieving a low calibration error at a narrow accuracy range but also analyze the performance of the method across different accuracy levels. We perform this analysis in Fig. \ref{fig:acc_ece}. We observe that (FL)Calib-1 performs the best in low accuracy ranges up to 50\%, and Calib-n+PS achieves the lowest ECE scores for accuracies between 50\% and 70\%.  LLM Prob.+PS works the best for high accuracy (>70\%) settings mainly because LLM probabilities are usually overconfident.

\begin{figure*}[ht]

    \centering
  
    \begin{subfigure}{0.195\textwidth} 
        \centering
        \includegraphics[width=1\linewidth]{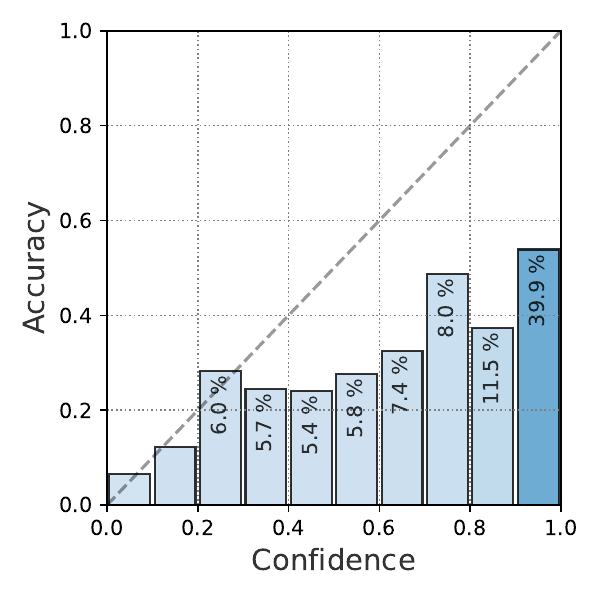}
        \caption{LLM Prob. }
    \end{subfigure}
    \begin{subfigure}{0.195\textwidth}
        \centering
        \includegraphics[width=1\linewidth]{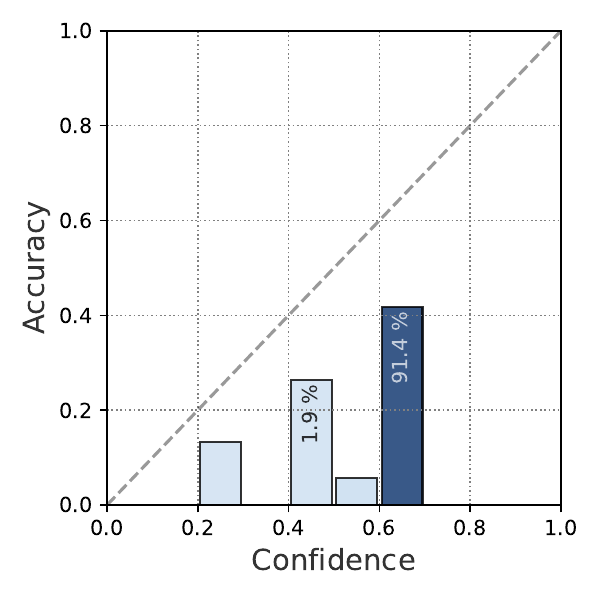}
         \caption{LLM Prob.+PS}
    \end{subfigure}
    \begin{subfigure}{0.195\textwidth} 
        \centering
        \includegraphics[width=1\linewidth]{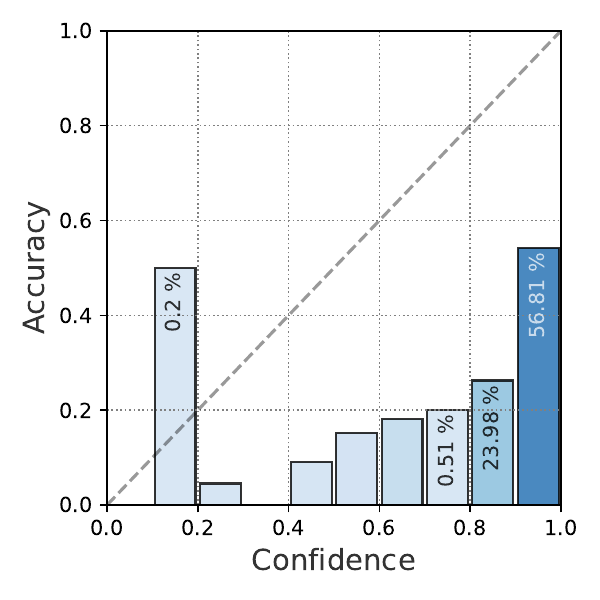}
        \caption{Verbalized \%}
    \end{subfigure}   
    \begin{subfigure}{0.195\textwidth}
        \centering
        \includegraphics[width=1\linewidth]{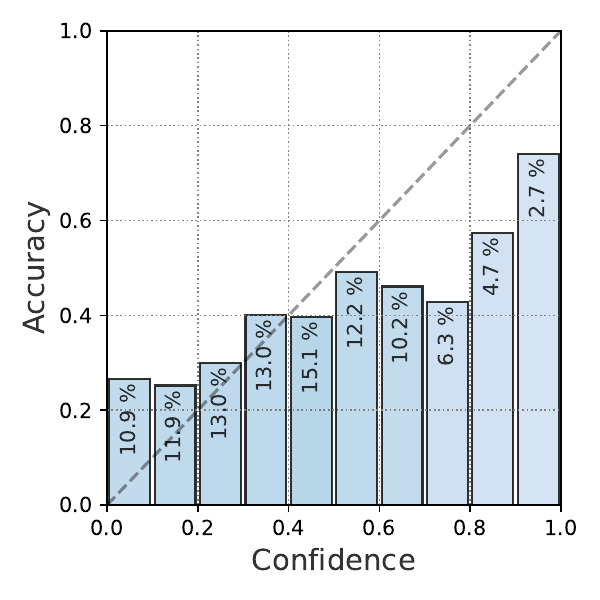}
         \caption{APRICOT}
    \end{subfigure}
    \begin{subfigure}{0.195\textwidth} 
        \centering
        \includegraphics[width=1\linewidth]{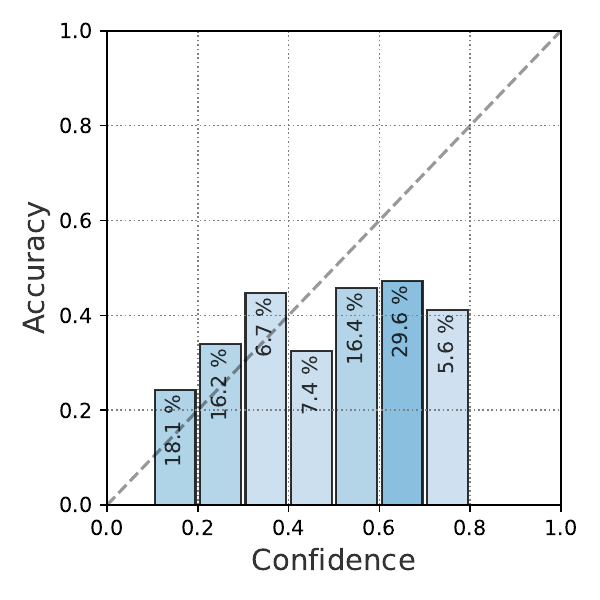}
        \caption{(BCE)Calib-1}
    \end{subfigure}
    
    \begin{subfigure}{0.195\textwidth}
        \centering
        \includegraphics[width=1\linewidth]{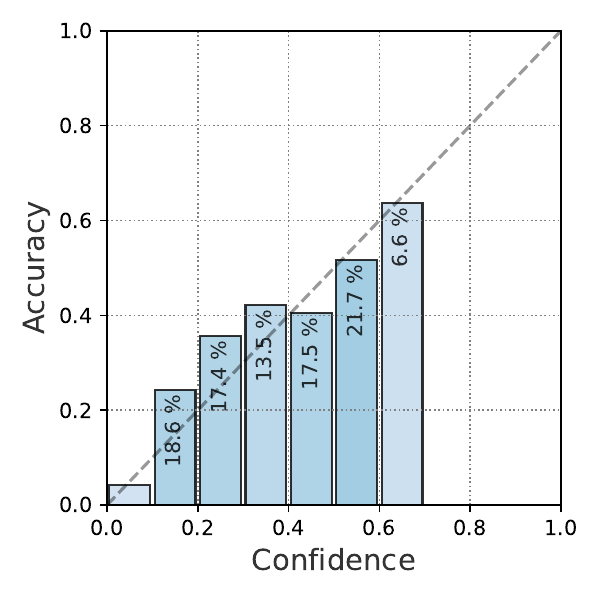}
         \caption{(BCE)Calib-n}
    \end{subfigure}
    \begin{subfigure}{0.195\textwidth} 
        \centering
        \includegraphics[width=1\linewidth]{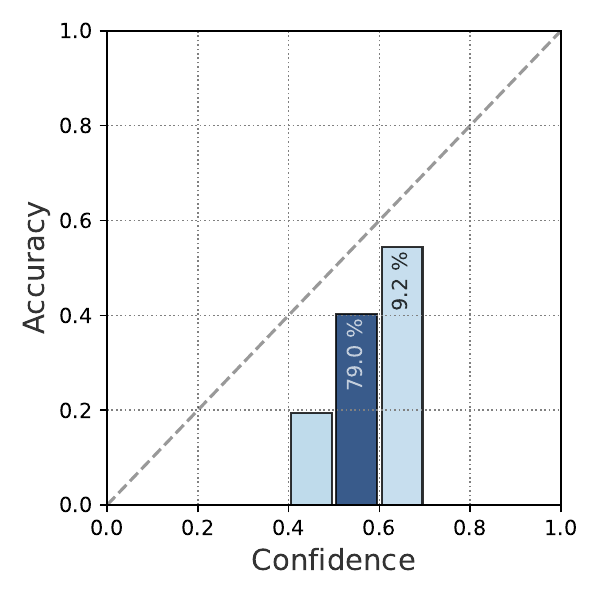}
        \caption{(BCE)Calib-n+PS}
    \end{subfigure}
    \begin{subfigure}{0.195\textwidth}
        \centering
        \includegraphics[width=1\linewidth]{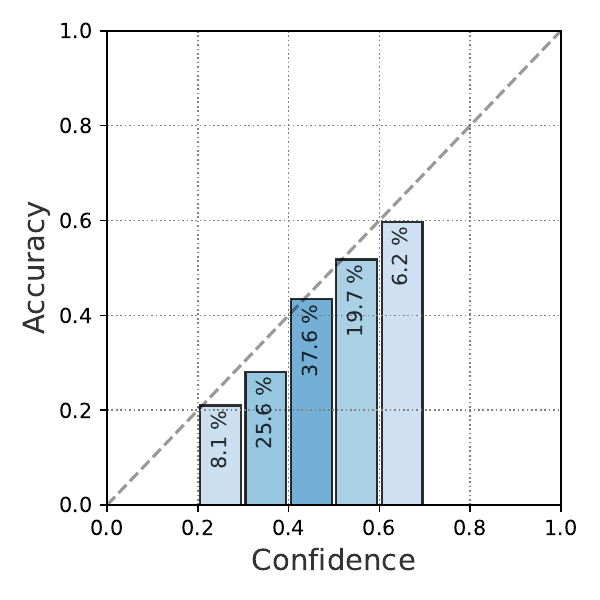}
         \caption{(FL)Calib-1}
    \end{subfigure}
    \begin{subfigure}{0.195\textwidth} 
        \centering
        \includegraphics[width=1\linewidth]{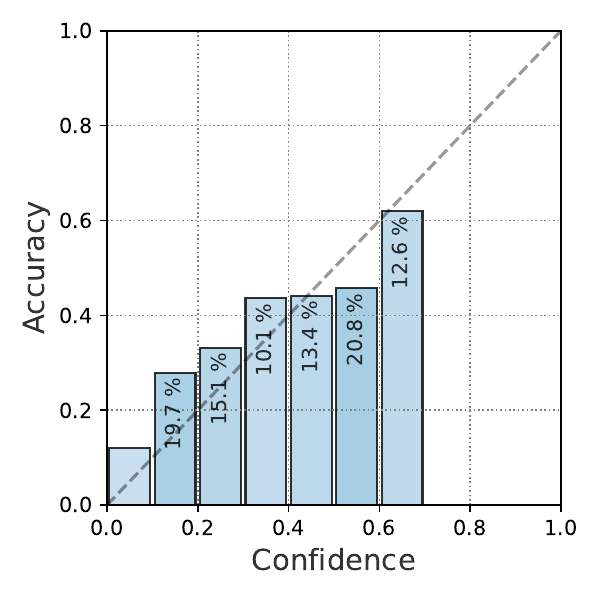}
        \caption{(FL)Calib-n}
    \end{subfigure}
    \begin{subfigure}{0.195\textwidth}
        \centering
        \includegraphics[width=1\linewidth]{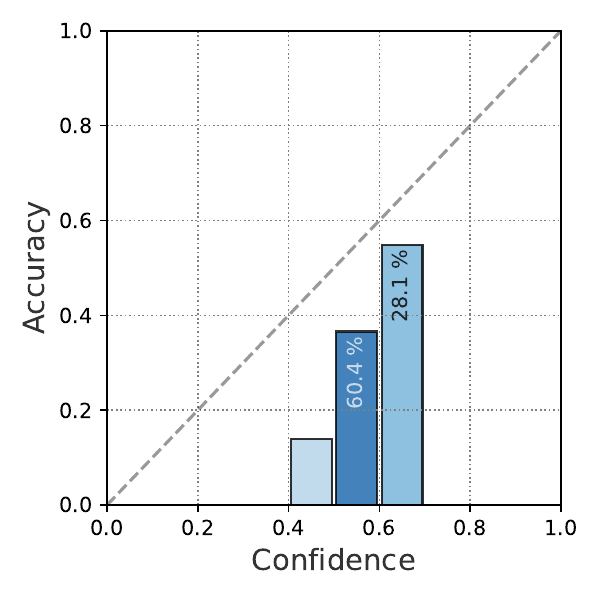}
         \caption{(FL)Calib-n+PS}
    \end{subfigure}
    \caption{Reliability diagrams for our different methods using 10 bins each for Llama3.1-70b on NQ. The color and the percentage number within each bar present the proportion of total data samples contained in each bin. More figures of other models and datasets are shown in Appendix \ref{other_diagrams}. }
    \label{fig:cali_diagram_plot}
\end{figure*}

\subsection{Detailed Fraction Results}

Table \ref{tb:nq_large} presents the test results of our methods and baselines on the NQ dataset when evaluated on five large-size LLMs. 
We analyze the performance of all methods across prompts, models, and evaluation metrics. The results reveal that the effectiveness of methods is highly influenced by the prompt styles and models. However, our proposed methods (Calib-*) achieve better calibration than baselines on most prompts and models. 

\textbf{Across Methods}:  LLM probabilities and Verbalized confidence show poor performance, exhibiting high ECE, ECE-t, and Brier scores across all prompts. Adding Platt Scaling (LLM Prob.+PS) improves calibration but is still outperformed by our proposed Calib-* methods. Among baselines, APRICOT performs better than Verbalized confidence, although it does not achieve the best results in most settings. Across all evaluation metrics, our Calib-* methods outperform others. For example, (BCE)Calib-n achieves the lowest ECE and Brier scores, particularly with the CoT and few-shot prompts, e.g., ECE of 0.049 on Llama3.1-70b. PS only helps decrease the ECE-t scores but can not enhance overall performance, this is because PS is similar to ECE-t which is a posthoc method for scaling probabilities.
\textbf{Across Prompts}: Calibration quality improves with few-shot and CoT prompts compared to the verb. and zero-shot prompts. For instance, (BCE)Calib-1 achieves an ECE-t of 0.045 with the CoT prompts compared to 0.068 with the verb. prompts on Gemma2-27b.
\textbf{Across Models}: Larger models (e.g., Qwen2-72b and Llama3.1-70b) generally exhibit better calibration when using Calib-* methods, reflecting their better alignment with calibration techniques. For instance, (FL)Calib-n+PS achieves an ECE of 0.071 on Qwen2-72b compared to 0.095 on smaller models like Mixtral-8x7b.

\textbf{Reliability diagrams analysis.} Fig. \ref{fig:cali_diagram_plot} shows that PS produces a narrow range of confidence scores, indicating limited diversity in the emitted confidence levels.  LLM probabilities and Verbalized confidence often exhibit overconfidence, even when applied to datasets with low accuracy (<40\%, reported in Fig. \ref{fig:all_accuracy} in Appendix). In contrast, our Calib-* methods show a more conservative approach, aligning their confidence levels more closely with the true accuracy of the model, reflecting improved calibration and reliability. 

\section{Conclusion}
Previous studies \cite{ulmer-etal-2024-calibrating, tian2023just} have assessed calibration methods within limited settings, overlooking their generalization across diverse model sizes and prompt styles. This study addresses this limitation by conducting experiments on 12 LLMs with parameters ranging from 2B-72B and four prompt styles. We also comprehensively analyze the influence of response agreement and loss functions in LLM calibration. Experimental results show that both response agreement and FL loss enhance calibration from baselines, with (FL)Calib-1 achieving the best performance. We also find that few-shot prompts improve LLM accuracy and calibration, and auxiliary-based methods show robust performance across diverse settings---maintaining stable calibration regardless of accuracy levels. These findings highlight the effectiveness of various calibration strategies and encourage future methods to re-evaluate the importance of our explored factors for achieving reliable confidence estimation.

\section{Limitations}

Although our work covers a wide range of factors, there are potentially more factors worth exploring. For example, we analyze a wide range of prompt types, including Few-shot and Chain-of-Thought, the influence of fine-grained prompt variations or automatically generated prompts remains unexplored. The interplay between prompt engineering and calibration could warrant deeper investigation.

Our study focuses on calibration performance metrics like ECE, ECE-t, and Brier scores. While these metrics are widely used, they may not fully capture all aspects of calibration quality, such as user-perceived confidence or task-specific utility.

Existing works use various ways of determining the accuracy of LLM generations. For instance, some works \cite{tian2023just, liu2024litcablightweightlanguagemodel} use LLMs as a Judge, other works \cite{ulmer-etal-2024-calibrating, xiong2024llms} use certain metrics such as extract match or ROUGE score. While the optimal solution is underexplored, we choose the more commonly used and cost-efficient method.

Future work can address these limitations by testing broader LLMs and tasks, automating prompt optimization, and developing hybrid approaches that adapt to varying accuracy levels and application constraints.

\section{Acknowledgments} 
This research has been funded by the Vienna Science and Technology Fund (WWTF)[10.47379/VRG19008] ”Knowledge infused Deep Learning for Natural Language Processing”.
\bibliography{custom}
\clearpage
\appendix

\section{Appendix}

\label{sec:appendix}

\subsection{Prompts for target LLM} \label{app:llm_prompts}
\subsubsection{Verb. prompt}
\begin{tcolorbox}[width=.48\textwidth, colframe=blue!20, colback=blue!10, sharp corners]
Provide your best guess 
and the probability that it is correct (0.0 to 1.0) for the following question. Give ONLY the guess and probability, no other words or explanation. For example: \textbackslash n\textbackslash n Guess: <most likely guess, as short as possible; not a complete sentence, just the guess!>\textbackslash n "Probability: <the probability between 0.0 and 1.0 that your guess is correct, without any extra commentary whatsoever; just the probability!>\textbackslash n\textbackslash n The question is: [Question]
\end{tcolorbox}

\subsubsection{Zero-shot prompt}

\begin{tcolorbox}[width=.48\textwidth, colframe=blue!20, colback=blue!10, sharp corners]
Provide your best guess for the following question. Give ONLY the guess, no other words or explanations, as short as possible; not a complete sentence, just the guess! \textbackslash n\textbackslash n The question is: [Question]
\end{tcolorbox}

\subsubsection{CoT prompt}
\begin{tcolorbox}[width=.48\textwidth, colframe=blue!20, colback=blue!10, sharp corners]
Briefly answer the following question by thinking step by step. Give the final answer (start with 'Answer: ' ) with minimal words at the end. \textbackslash n\textbackslash n The question is: [Question]
\end{tcolorbox}

\subsubsection{Few-shot prompt}

\begin{tcolorbox}[width=.48\textwidth, colframe=blue!20, colback=blue!10, sharp corners]
user: [Question 1]  ~~  assistant: [Target 1] \\
user: [Question 2]  ~~  assistant: [Target 2] \\
...\\
user: [Question 6]  ~~  assistant: [Target 6] \\
user: [Question]   ~~  assistant:
\end{tcolorbox}

\subsection{Prompt for Judge}\label{app:judge_prompts}
\begin{tcolorbox}[width=.48\textwidth, colframe=blue!20, colback=blue!10, sharp corners]
Task Description: \textbackslash n 
An instruction (might include an Input inside it), a response to evaluate, a reference answer that gets a score of 1, and a score rubric representing a evaluation criteria are given. 

1. Write detailed feedback that assesses the quality of the response strictly based on the given score rubric, not evaluating in general.

2. After writing feedback, write a score that is an integer between 0 and 1. You should refer to the score rubric.

3. The output format should look as follows: "Feedback: (write a feedback for criteria) [RESULT] (an integer number between 0 and 1)"

4. Please do not generate any other opening, closing, and explanations.

The instruction to evaluate:[Question]

Response to evaluate: [LLM Answer]

Reference Answer (Score 1): [Target]

Score Rubrics:\textbackslash n 
Score 0: the response and reference answer to the instruction are not semantically equivalent.\textbackslash n 
Score 1: the response and reference answer to the instruction are semantically equivalent.

Feedback:

\end{tcolorbox}

\subsection{Technical Details}
After a grid search of hyperparameters, we trained our auxiliary models (BERT-base, 110M parameters) using a learning rate of 1e-5 and a batch size of 16 for five epochs.  All experiments, including LLM inferences, are performed on a maximum of 2 NVIDIA H100 GPUs. The training time for one epoch of 2k samples is around 200 seconds on one GPU, this time can be different depends on the dataset size and number of joint LLMs in training.

\subsection{Aggregated Results Across Different Configurations}

\begin{figure*}[ht!]
    \centering
    \begin{subfigure}{0.48\textwidth} 
        \centering
        \includegraphics[width=1\linewidth]{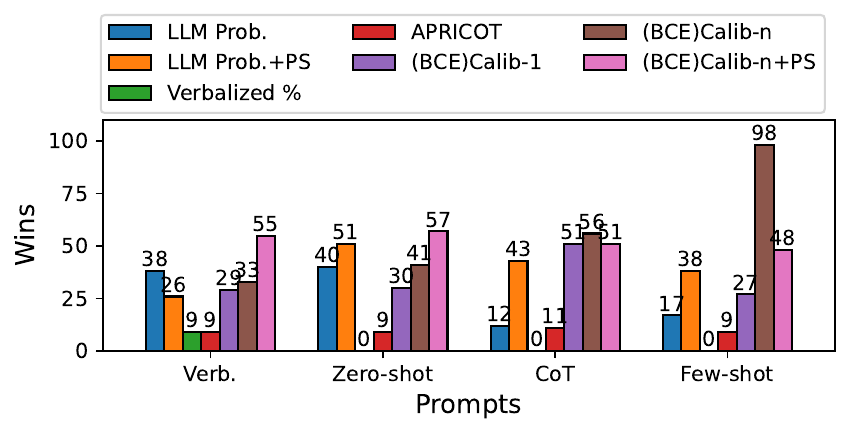}
    \end{subfigure}
    \hspace{.3cm} 
    \begin{subfigure}{0.48\textwidth}
        \centering
        \includegraphics[width=1\linewidth]{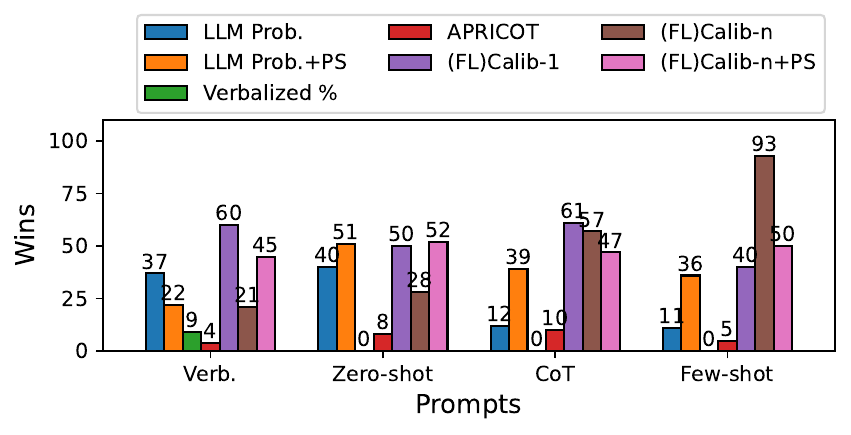}
    \end{subfigure}

    \begin{subfigure}{0.45\textwidth} 
        \centering
        \includegraphics[width=1\linewidth]{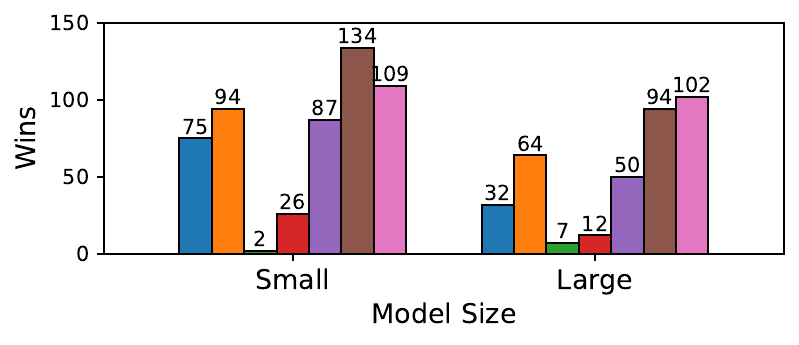}
    \end{subfigure}
    \hspace{.3cm}
    \begin{subfigure}{0.45\textwidth}
        \centering
        \includegraphics[width=1\linewidth]{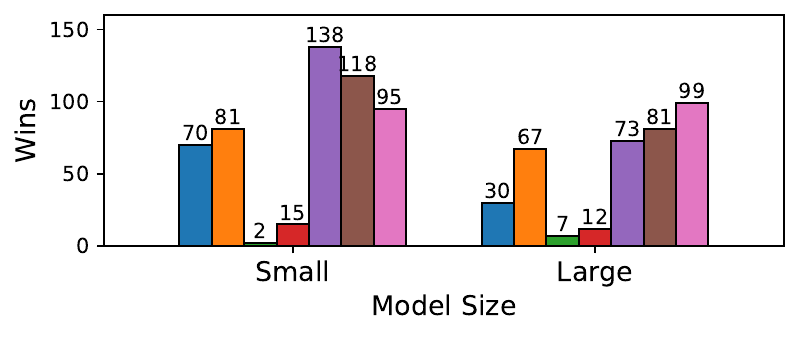}
    \end{subfigure}
    \begin{subfigure}{0.48\textwidth} 
        \centering
        \includegraphics[width=1\linewidth]{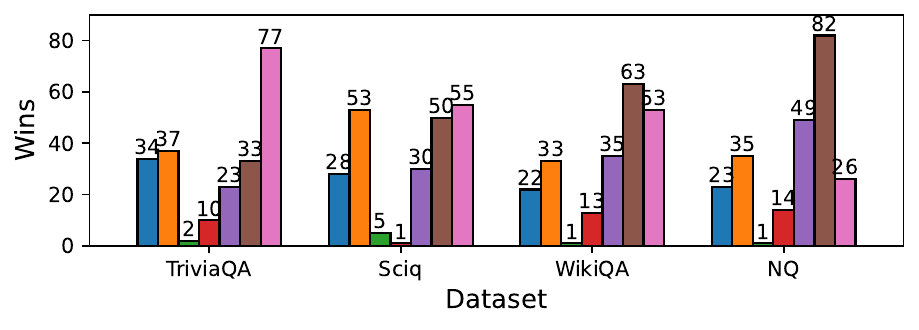}
    \end{subfigure}
    \hspace{.3cm} 
    \begin{subfigure}{0.48\textwidth}
        \centering
        \includegraphics[width=1\linewidth]{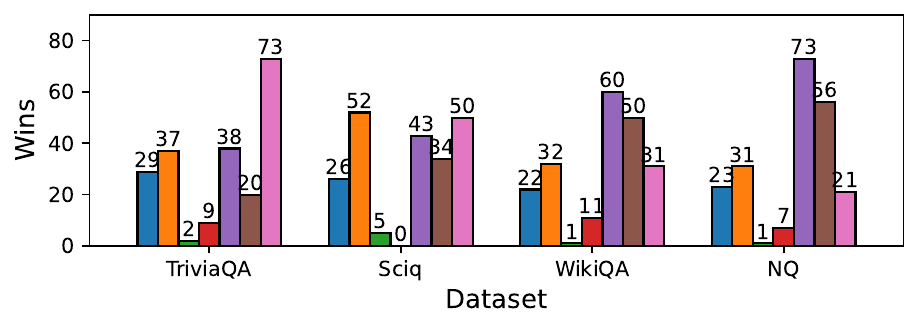}
    \end{subfigure}

    \caption{Performance Comparison results of different methods against baselines across three setting configurations (prompt styles, model sizes and datasets).}
    \label{fig:different_setting}
\end{figure*}

Fig. \ref{fig:different_setting} shows the calibration performances of different methods across different configurations such as prompt styles, model sizes and datasets.
We observe that the best method is highly dependent on these factors.

\textbf{Prompt specific performance.}
The first row of Fig. \ref{fig:different_setting} presents the prompt specific performance for different methods. We observe that Calib-n always outperforms Calib-1 across different prompt styles when both are optimized with BCE loss. When FL loss is applied, Calib-n only outperforms Calib-1 in few-shot prompts. We hypothesize that the longer LLM answers generated with few-shot prompts usually lead to higher response agreement and thus enhance the performance of Calib-n.

\textbf{Model size specific performance.}
The second row of Fig. \ref{fig:different_setting} presents the model-size specific performance for different methods. We find that Calib-n outperforms Calib-1 on large-size models. However, (FL)Calib-1 performs better for small-size models. We also observe that the best method is model size dependent.

\textbf{Dataset specific performance.}
The second row of Fig. \ref{fig:different_setting} presents the dataset specific performance for different methods. (BCE)Calib-n consistently achieves better performances than (BCE)Calib-1 over all datasets. In contrast, (FL)Calib-1 always outperform (FL)Calib-n. 

\subsection{Accuracy Statistics of LLMs} \label{model_accuracy}
\begin{figure*}
    \centering
    \includegraphics[width=1\linewidth]{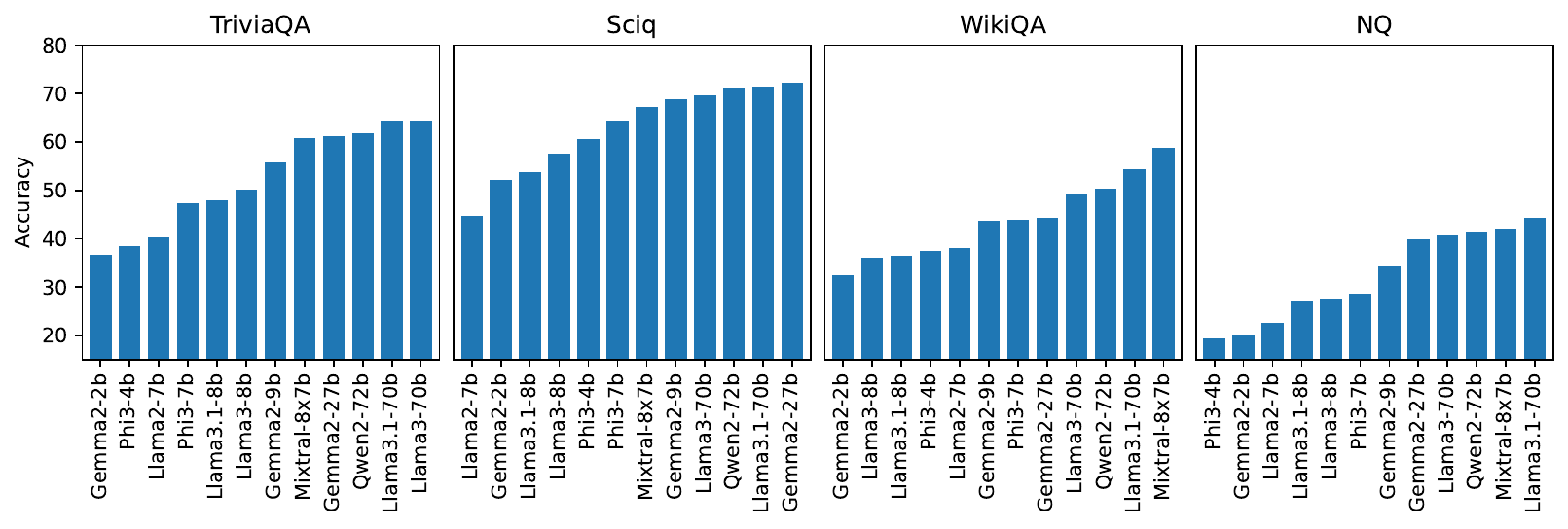}
    \caption{Model performance ranking across different datasets, performance is averaged over four prompt styles.}
    \label{fig:accuracy_rank}
\end{figure*}

\begin{figure*}
    \centering
    \includegraphics[width=1\linewidth]{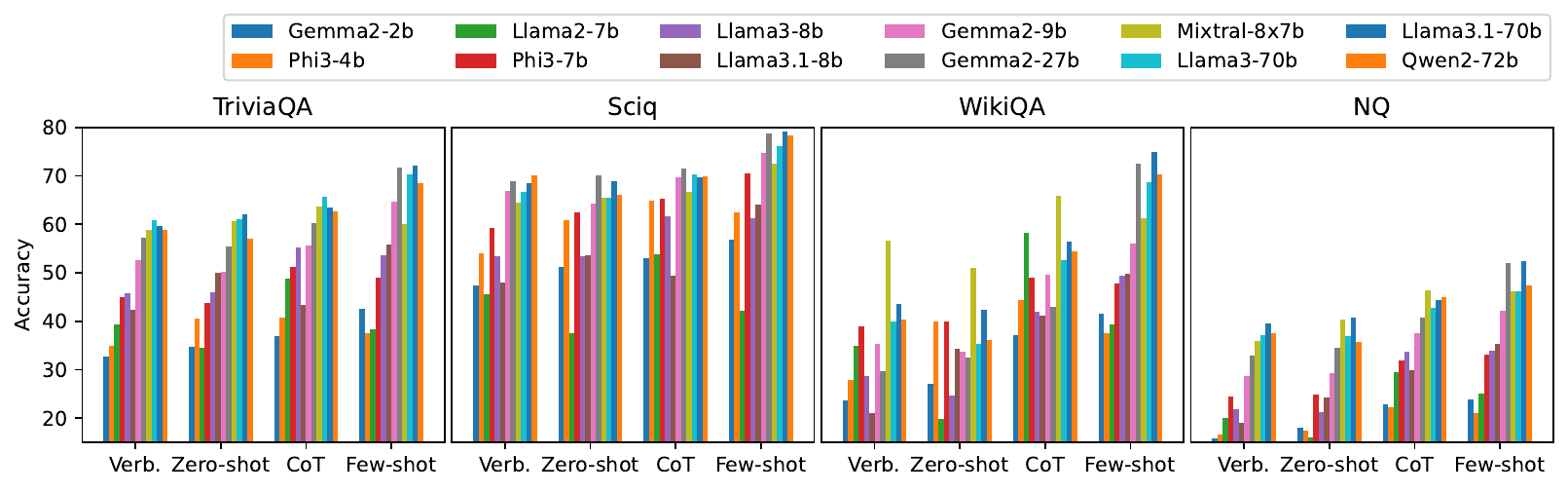}
    \caption{Model performance across different prompts and datasets}
    \label{fig:all_accuracy}
\end{figure*}

We present Fig. \ref{fig:accuracy_rank} and \ref{fig:all_accuracy} to demonstrate the accuracy performance of LLMs across different prompts and datasets.  Large size models typically yield better performance than small size models. Few shot prompts improve the performance more than other prompt styles.  Most LLMs achieve their highest accuracy on the Sciq dataset, while the NQ dataset proves to be the most challenging. 
\subsection{Reliability Diagrams}\label{other_diagrams}
\begin{figure*}[ht!]

    \centering
  
    \begin{subfigure}{0.161\textwidth} 
        \centering
        \includegraphics[width=1\linewidth]{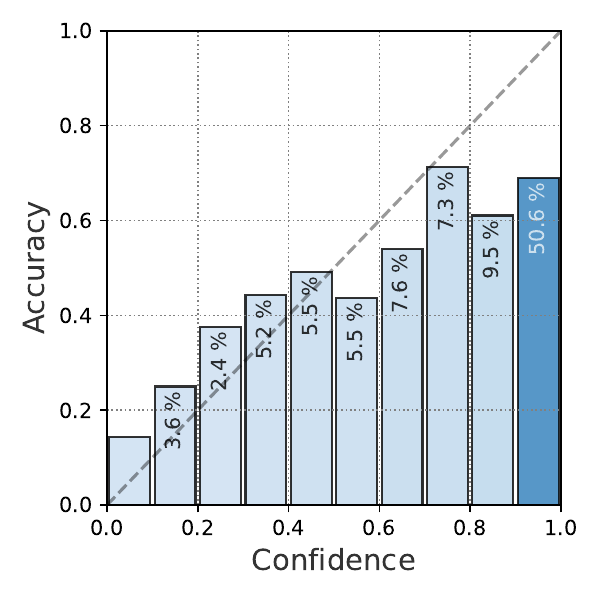}
        \caption{\scriptsize LLM Prob. }
    \end{subfigure}
    \begin{subfigure}{0.161\textwidth}
        \centering
        \includegraphics[width=1\linewidth]{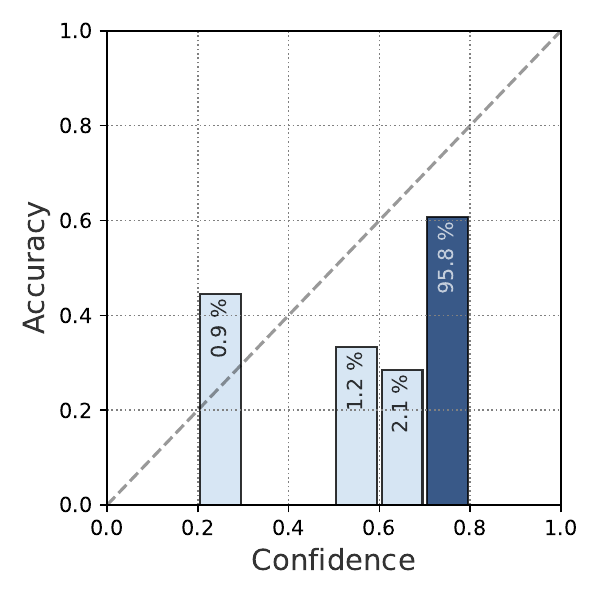}
         \caption{\scriptsize LLM Prob.+PS}
    \end{subfigure}
    \begin{subfigure}{0.161\textwidth} 
        \centering
        \includegraphics[width=1\linewidth]{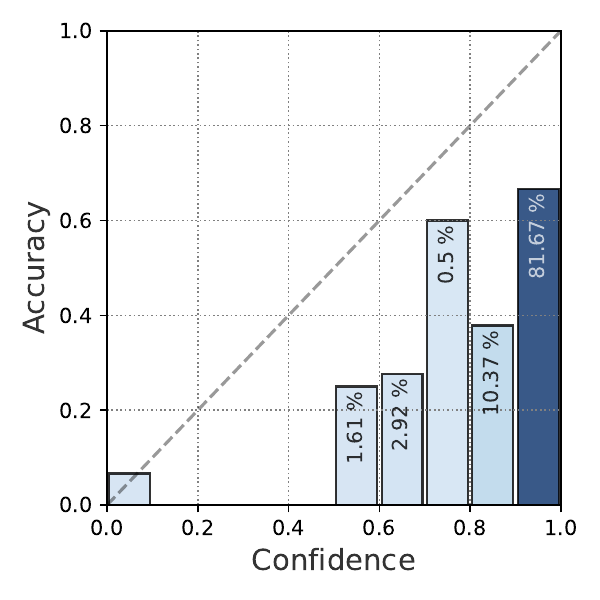}
        \caption{\scriptsize Verbalized \%}
    \end{subfigure}   
    \begin{subfigure}{0.161\textwidth}
        \centering
        \includegraphics[width=1\linewidth]{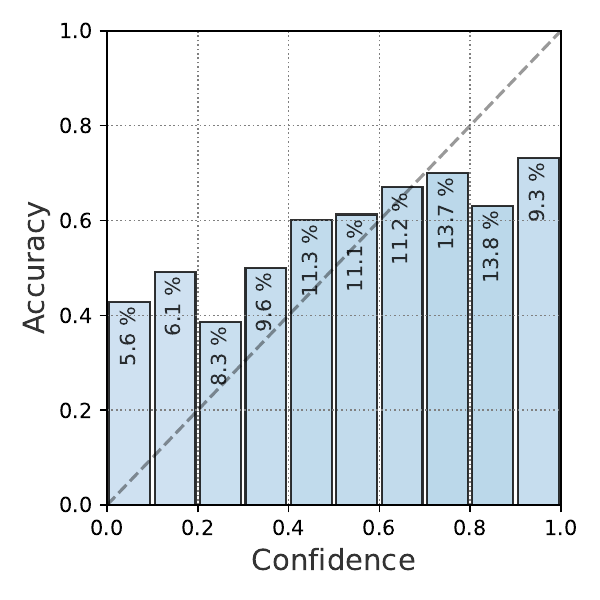}
         \caption{\scriptsize APRICOT}
    \end{subfigure}
    \begin{subfigure}{0.161\textwidth} 
        \centering
        \includegraphics[width=1\linewidth]{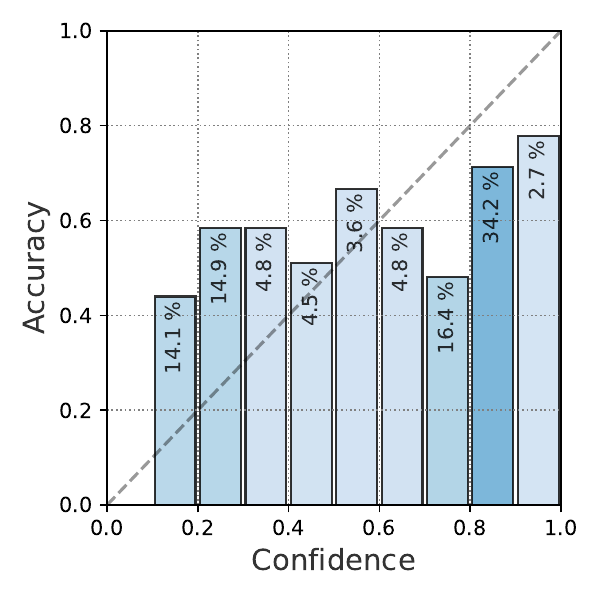}
        \caption{\scriptsize (AUC)Calib-1}
    \end{subfigure}
    \begin{subfigure}{0.161\textwidth}
        \centering
        \includegraphics[width=1\linewidth]{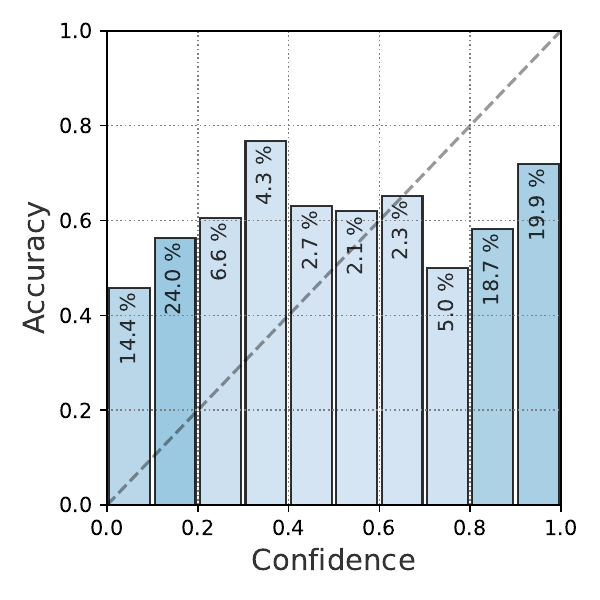}
         \caption{\scriptsize AUC)Calib-n}
    \end{subfigure}
    
    \begin{subfigure}{0.161\textwidth} 
        \centering
        \includegraphics[width=1\linewidth]{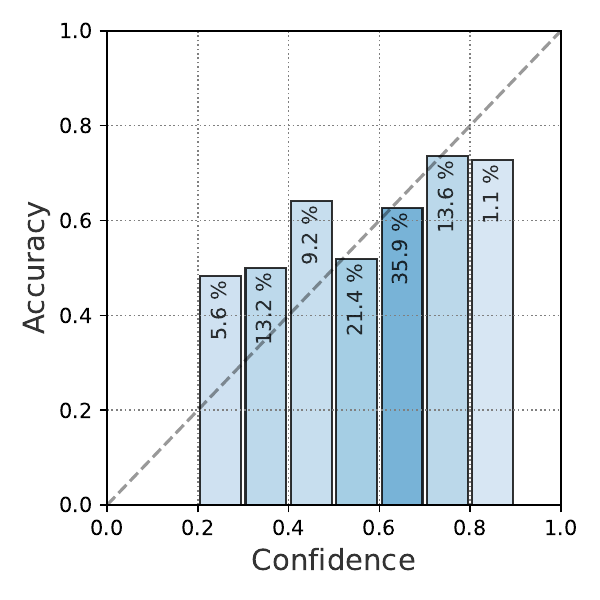}
        \caption{\scriptsize (BCE)Calib-1}
    \end{subfigure}
    \begin{subfigure}{0.161\textwidth}
        \centering
        \includegraphics[width=1\linewidth]{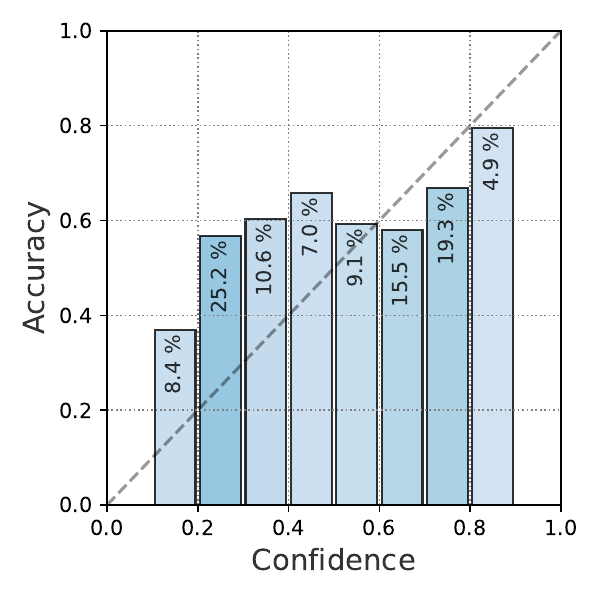}
         \caption{\scriptsize (BCE)Calib-n}
    \end{subfigure}
    \begin{subfigure}{0.161\textwidth} 
        \centering
        \includegraphics[width=1\linewidth]{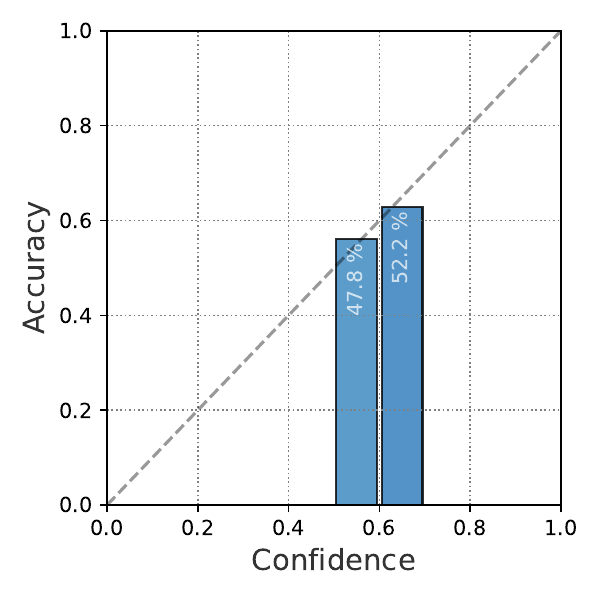}
        \caption{\scriptsize (BCE)Calib-n+PS}
    \end{subfigure}
    \begin{subfigure}{0.161\textwidth}
        \centering
        \includegraphics[width=1\linewidth]{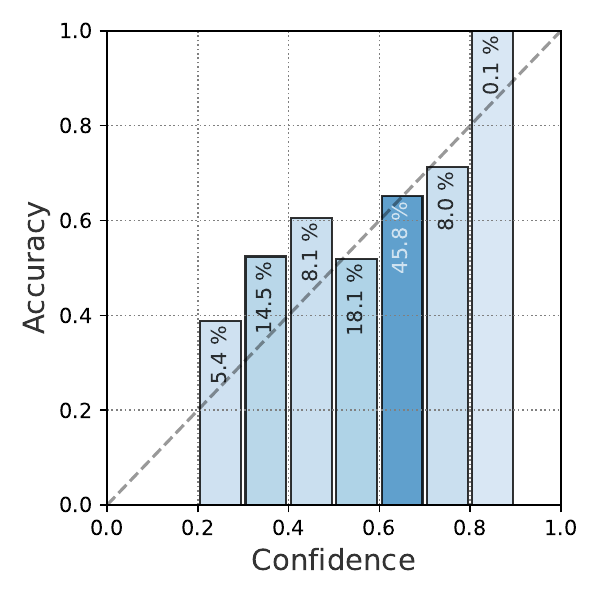}
         \caption{\scriptsize (FL)Calib-1}
    \end{subfigure}
    \begin{subfigure}{0.161\textwidth} 
        \centering
        \includegraphics[width=1\linewidth]{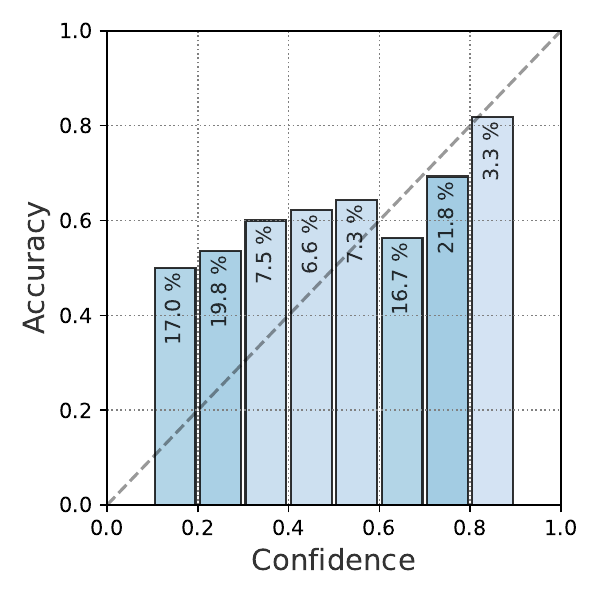}
        \caption{\scriptsize (FL)Calib-n}
    \end{subfigure}
    \begin{subfigure}{0.161\textwidth}
        \centering
        \includegraphics[width=1\linewidth]{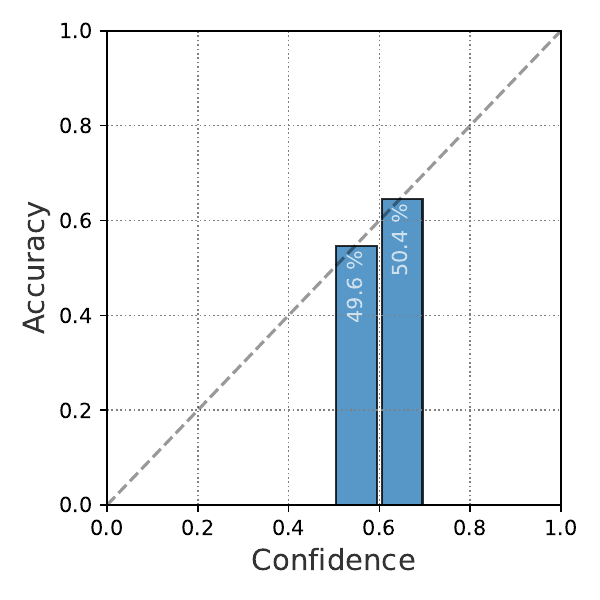}
         \caption{\scriptsize (FL)Calib-n+PS}
    \end{subfigure}
    \caption{Reliability diagrams for Llama3.1-70b on TriviaQA with Verb. prompts. }
    \label{fig:cali_diagram_triviaqa}
\end{figure*}

\begin{figure*}[htbp]

    \centering
  
    \begin{subfigure}{0.161\textwidth} 
        \centering
        \includegraphics[width=1\linewidth]{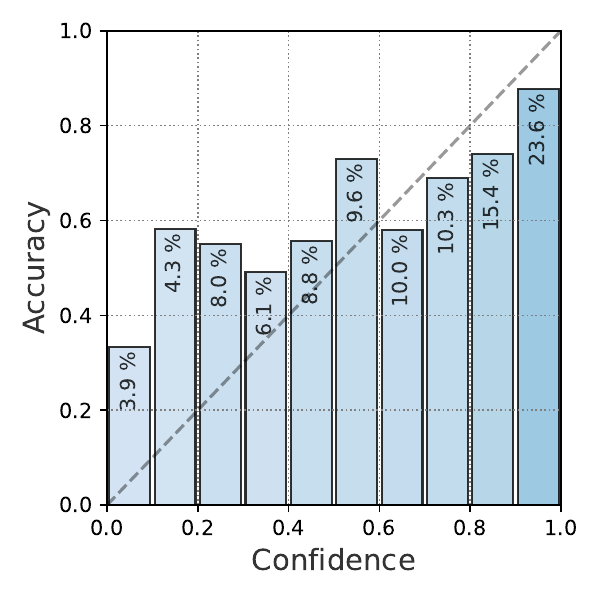}
        \caption{\scriptsize LLM Prob. }
    \end{subfigure}
    \begin{subfigure}{0.161\textwidth}
        \centering
        \includegraphics[width=1\linewidth]{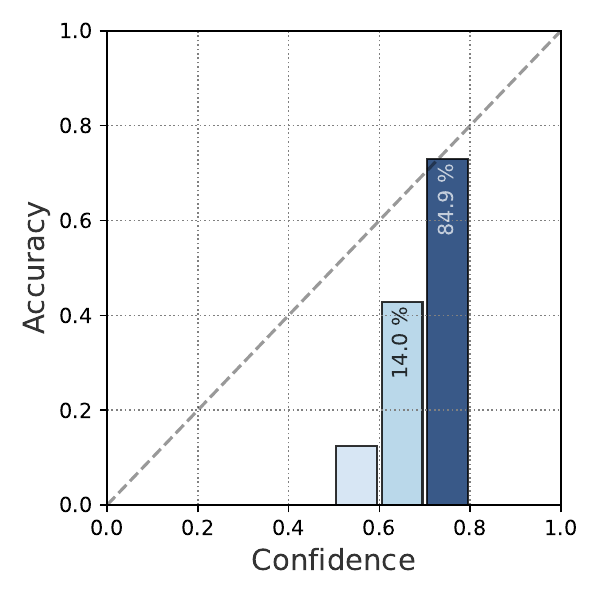}
         \caption{\scriptsize LLM Prob.+PS}
    \end{subfigure}
    \begin{subfigure}{0.161\textwidth} 
        \centering
        \includegraphics[width=1\linewidth]{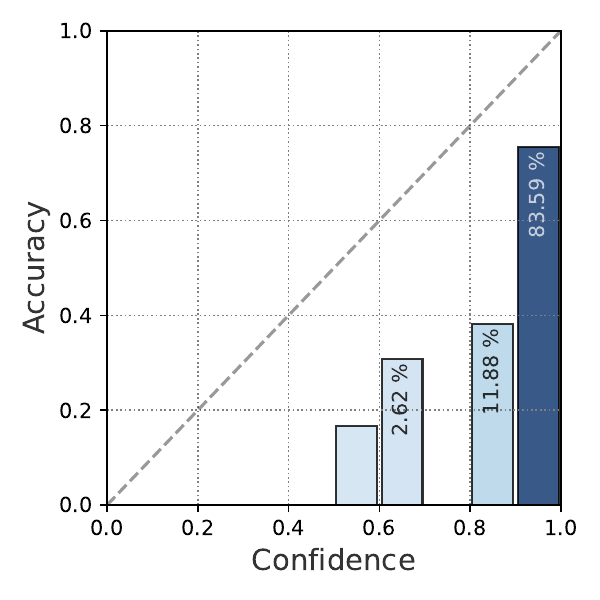}
        \caption{\scriptsize Verbalized \%}
    \end{subfigure}   
    \begin{subfigure}{0.161\textwidth}
        \centering
        \includegraphics[width=1\linewidth]{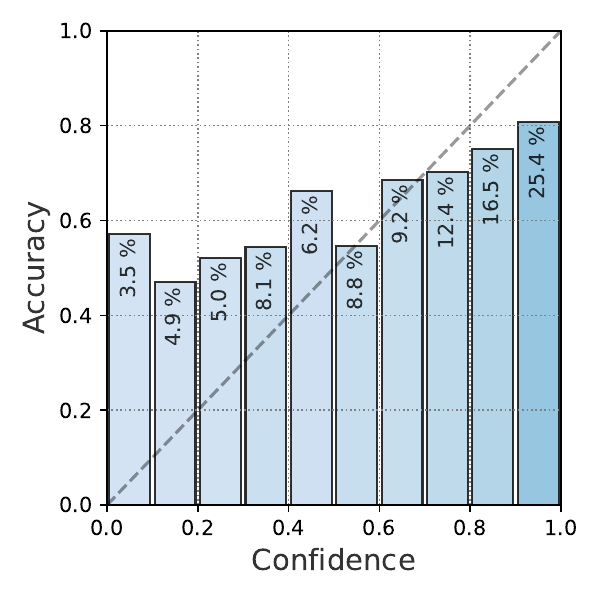}
         \caption{\scriptsize APRICOT}
    \end{subfigure}
    \begin{subfigure}{0.161\textwidth} 
        \centering
        \includegraphics[width=1\linewidth]{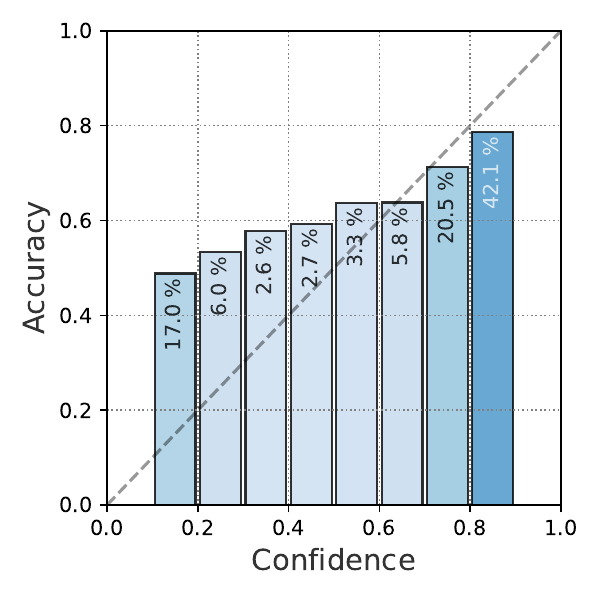}
        \caption{\scriptsize (AUC)Calib-1}
    \end{subfigure}
    \begin{subfigure}{0.161\textwidth}
        \centering
        \includegraphics[width=1\linewidth]{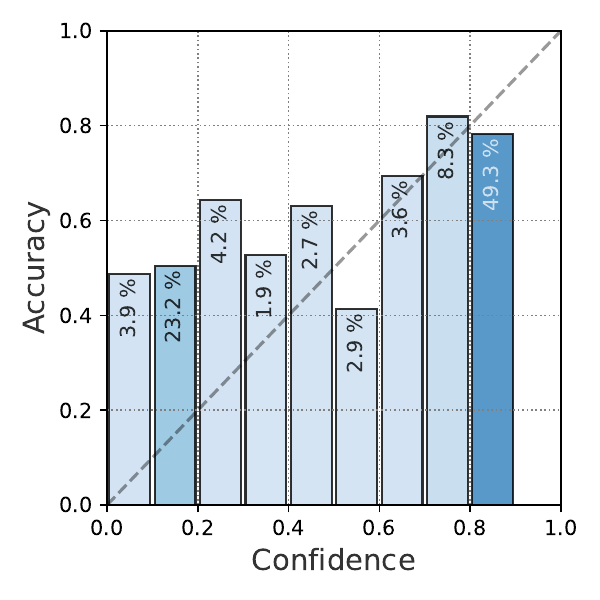}
         \caption{\scriptsize AUC)Calib-n}
    \end{subfigure}
    
    \begin{subfigure}{0.161\textwidth} 
        \centering
        \includegraphics[width=1\linewidth]{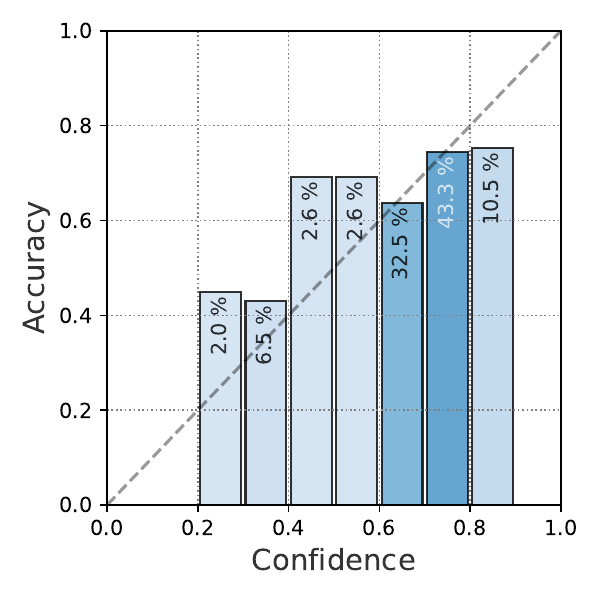}
        \caption{\scriptsize (BCE)Calib-1}
    \end{subfigure}
    \begin{subfigure}{0.161\textwidth}
        \centering
        \includegraphics[width=1\linewidth]{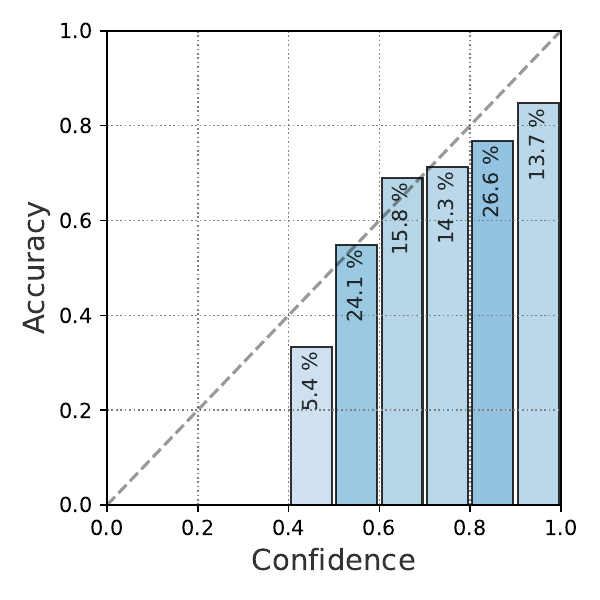}
         \caption{\scriptsize (BCE)Calib-n}
    \end{subfigure}
    \begin{subfigure}{0.161\textwidth} 
        \centering
        \includegraphics[width=1\linewidth]{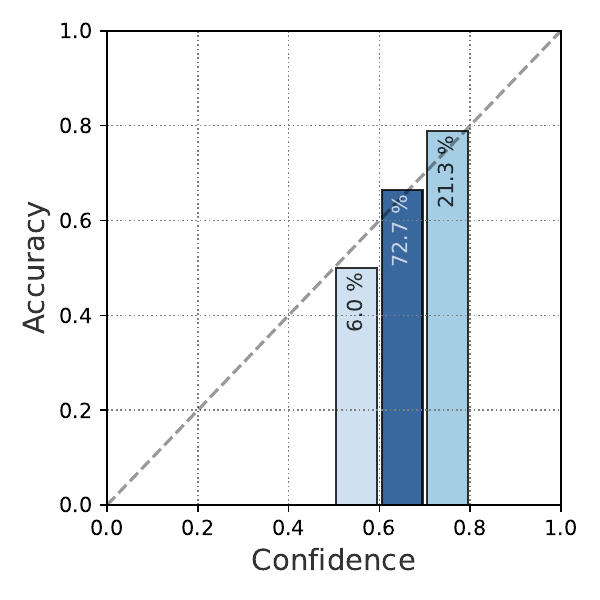}
        \caption{\scriptsize (BCE)Calib-n+PS}
    \end{subfigure}
    \begin{subfigure}{0.161\textwidth}
        \centering
        \includegraphics[width=1\linewidth]{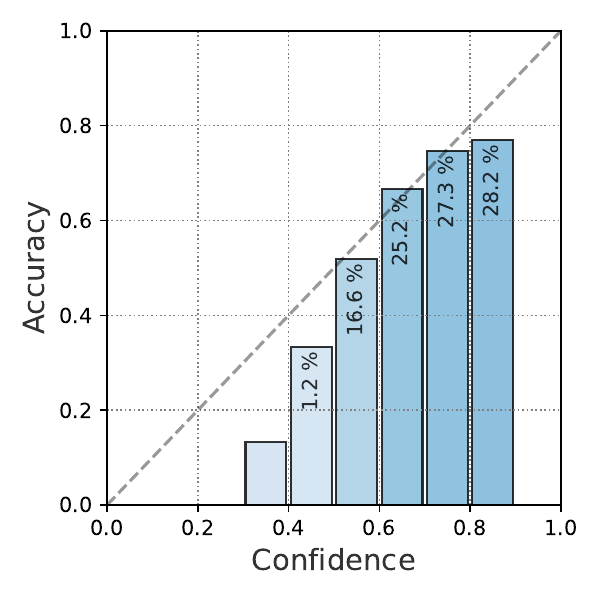}
         \caption{\scriptsize (FL)Calib-1}
    \end{subfigure}
    \begin{subfigure}{0.161\textwidth} 
        \centering
        \includegraphics[width=1\linewidth]{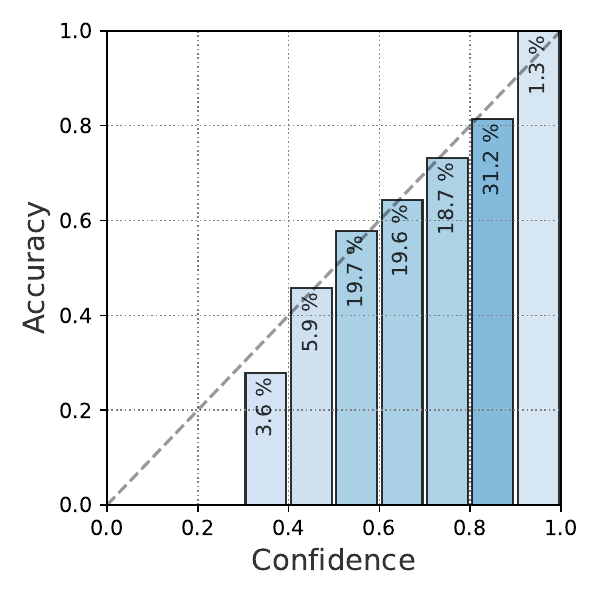}
        \caption{\scriptsize (FL)Calib-n}
    \end{subfigure}
    \begin{subfigure}{0.161\textwidth}
        \centering
        \includegraphics[width=1\linewidth]{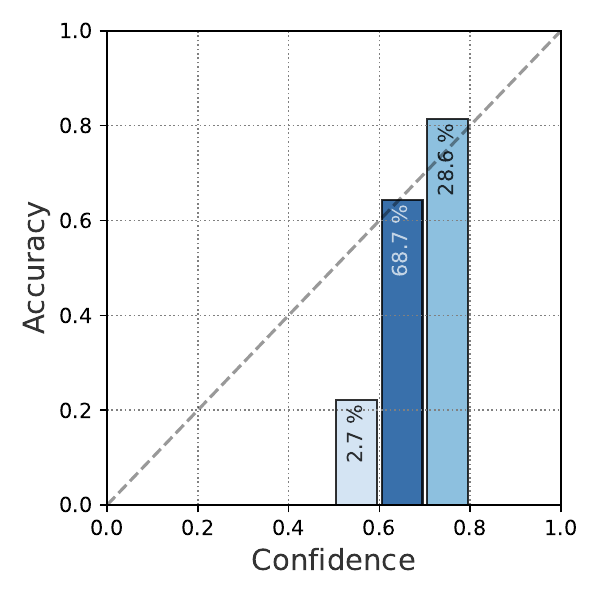}
         \caption{\scriptsize (FL)Calib-n+PS}
    \end{subfigure}
    \caption{Reliability diagrams for Llama3.1-70b on Sciq with Verb. prompts. }
    \label{fig:cali_diagram_plot_sciq}
\end{figure*}

\begin{figure*}[htbp]

    \centering
  
    \begin{subfigure}{0.161\textwidth} 
        \centering
        \includegraphics[width=1\linewidth]{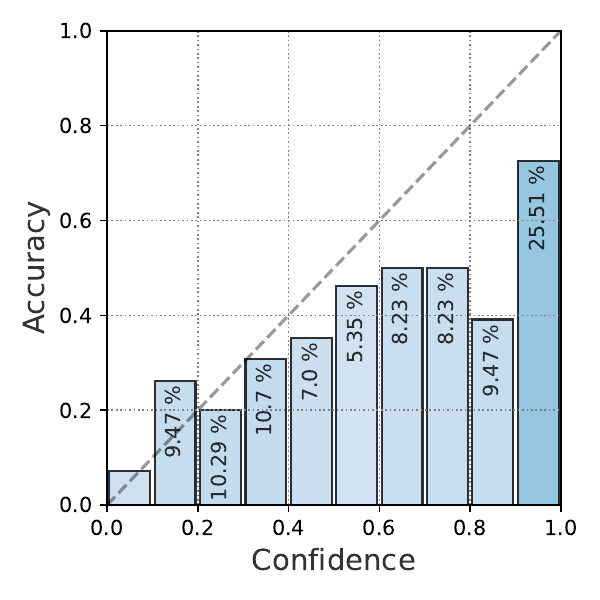}
        \caption{\scriptsize LLM Prob. }
    \end{subfigure}
    \begin{subfigure}{0.161\textwidth}
        \centering
        \includegraphics[width=1\linewidth]{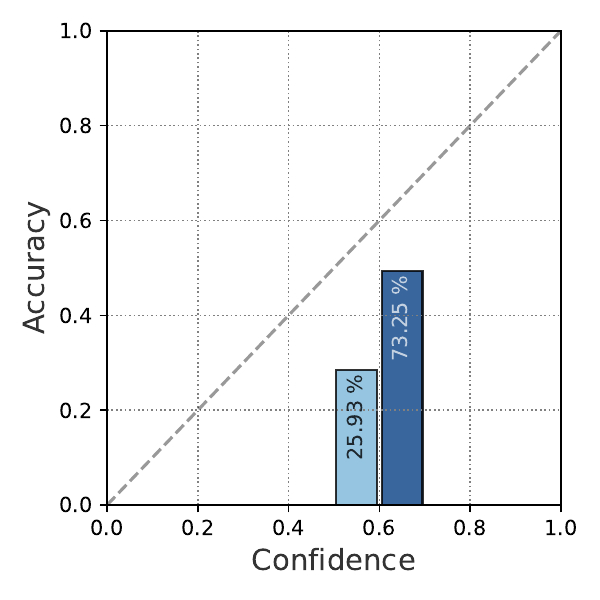}
         \caption{\scriptsize LLM Prob.+PS}
    \end{subfigure}
    \begin{subfigure}{0.161\textwidth} 
        \centering
        \includegraphics[width=1\linewidth]{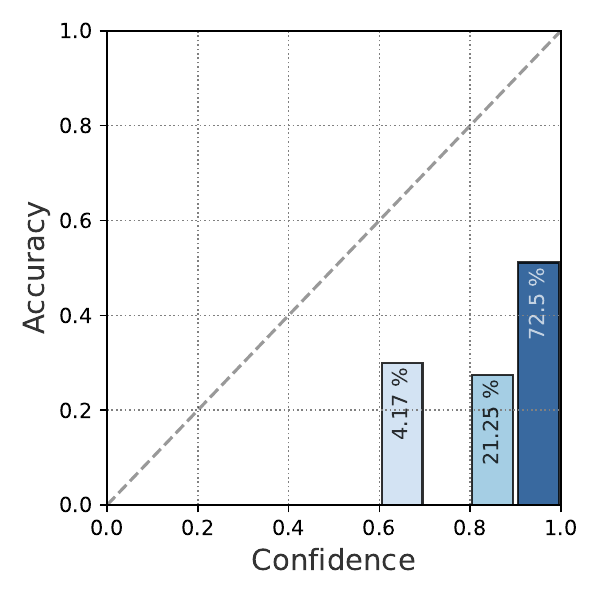}
        \caption{\scriptsize Verbalized \%}
    \end{subfigure}   
    \begin{subfigure}{0.161\textwidth}
        \centering
        \includegraphics[width=1\linewidth]{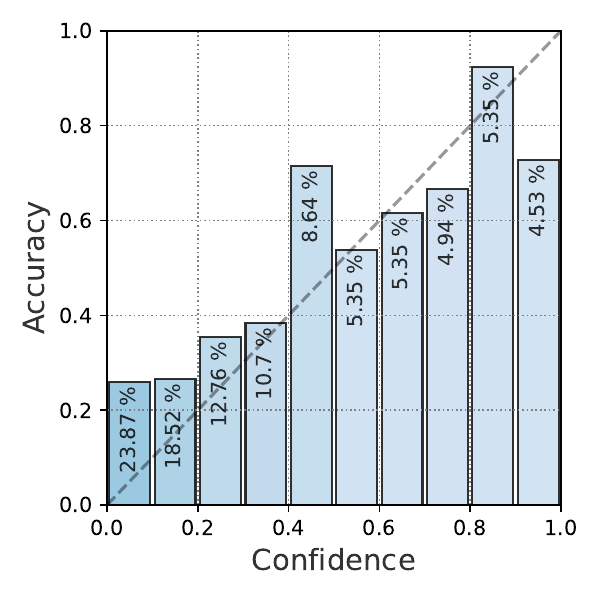}
         \caption{\scriptsize APRICOT}
    \end{subfigure}
    \begin{subfigure}{0.161\textwidth} 
        \centering
        \includegraphics[width=1\linewidth]{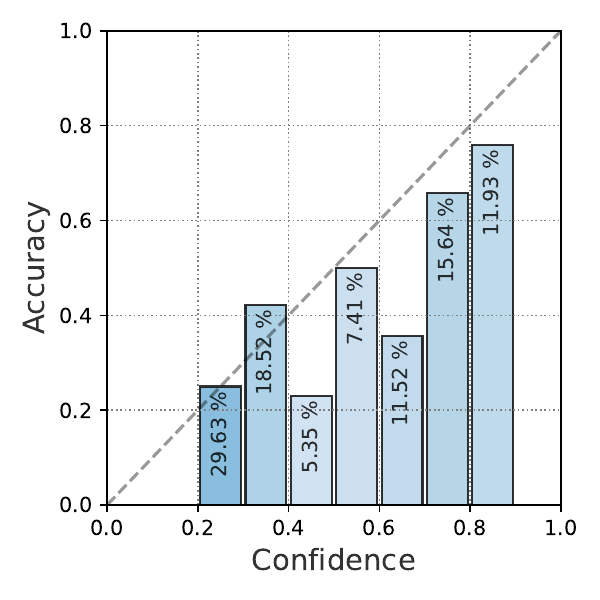}
        \caption{\scriptsize (AUC)Calib-1}
    \end{subfigure}
    \begin{subfigure}{0.161\textwidth}
        \centering
        \includegraphics[width=1\linewidth]{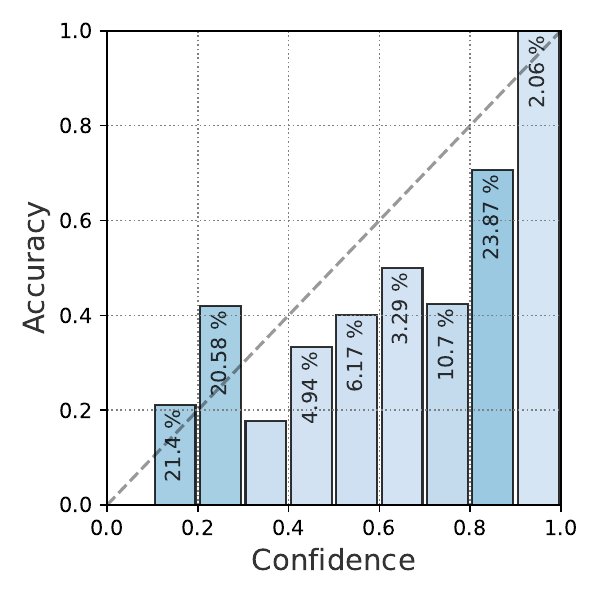}
         \caption{\scriptsize AUC)Calib-n}
    \end{subfigure}
    
    \begin{subfigure}{0.161\textwidth} 
        \centering
        \includegraphics[width=1\linewidth]{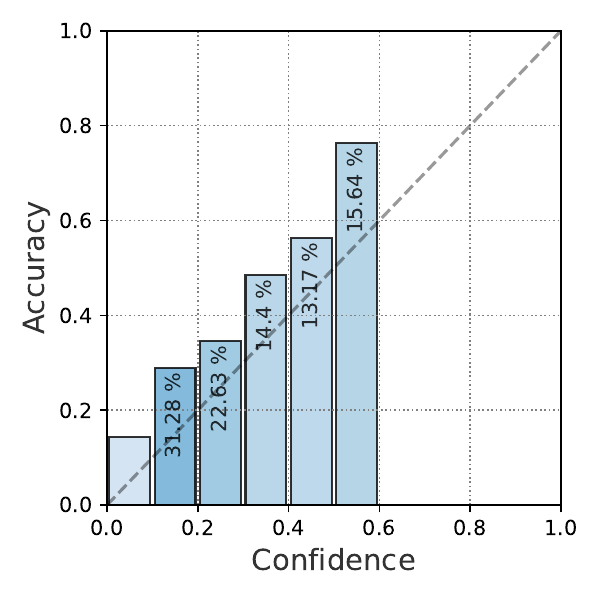}
        \caption{\scriptsize (BCE)Calib-1}
    \end{subfigure}
    \begin{subfigure}{0.161\textwidth}
        \centering
        \includegraphics[width=1\linewidth]{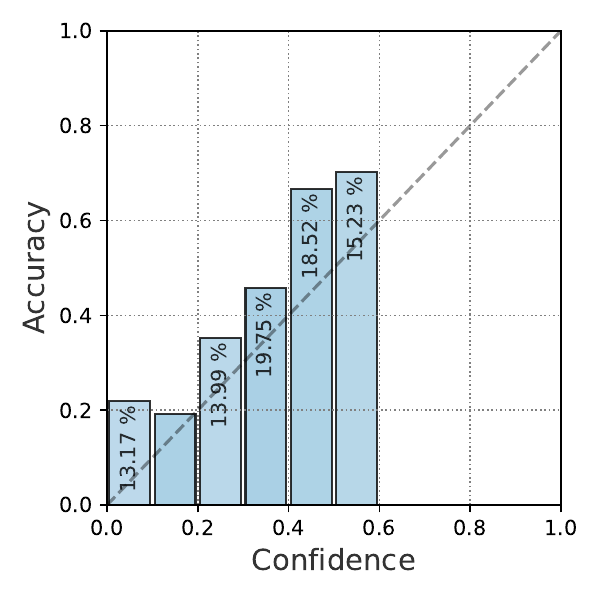}
         \caption{\scriptsize (BCE)Calib-n}
    \end{subfigure}
    \begin{subfigure}{0.161\textwidth} 
        \centering
        \includegraphics[width=1\linewidth]{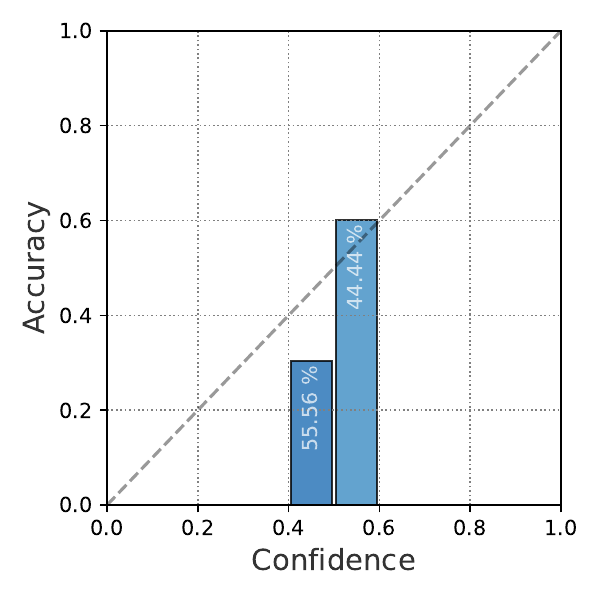}
        \caption{\scriptsize (BCE)Calib-n+PS}
    \end{subfigure}
    \begin{subfigure}{0.161\textwidth}
        \centering
        \includegraphics[width=1\linewidth]{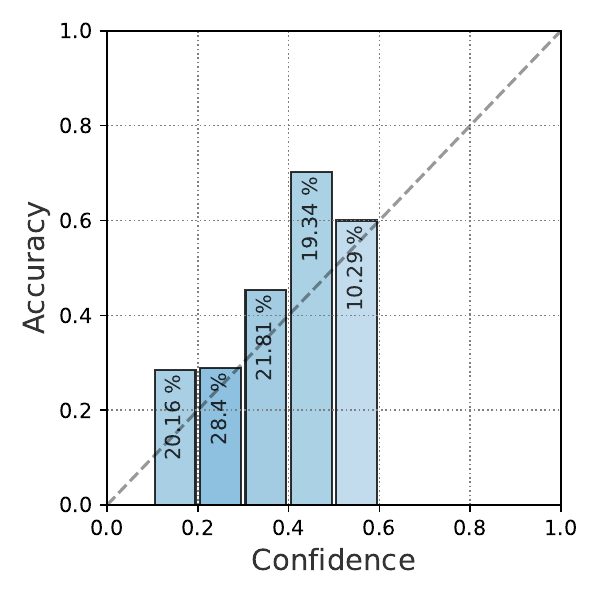}
         \caption{\scriptsize (FL)Calib-1}
    \end{subfigure}
    \begin{subfigure}{0.161\textwidth} 
        \centering
        \includegraphics[width=1\linewidth]{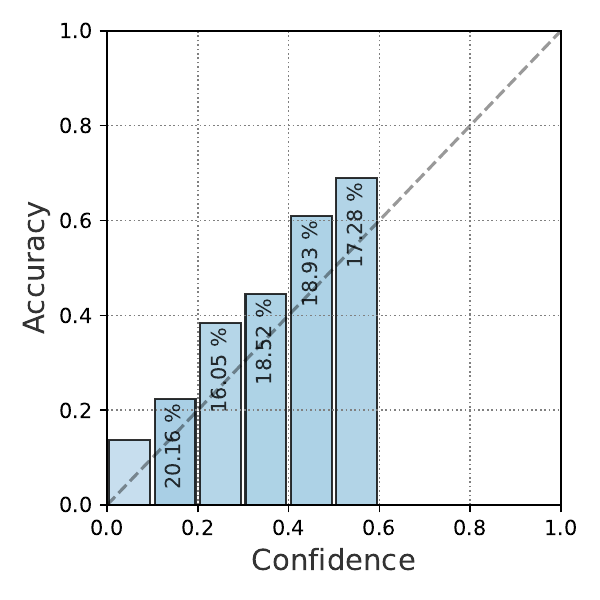}
        \caption{\scriptsize (FL)Calib-n}
    \end{subfigure}
    \begin{subfigure}{0.161\textwidth}
        \centering
        \includegraphics[width=1\linewidth]{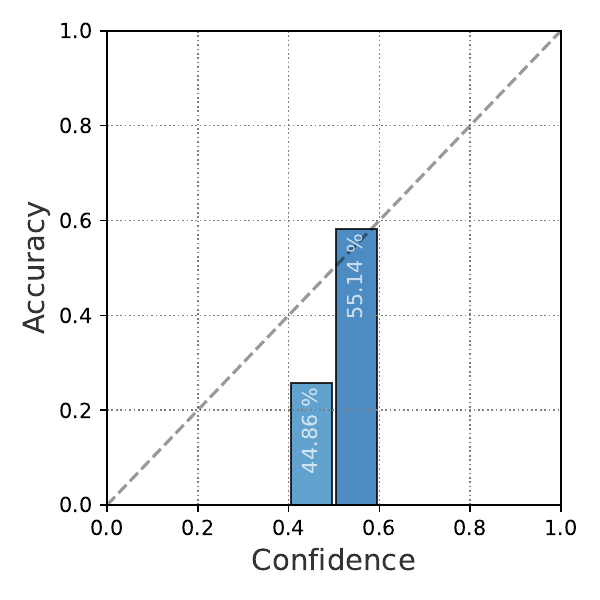}
         \caption{\scriptsize (FL)Calib-n+PS}
    \end{subfigure}
    \caption{Reliability diagrams for Llama3.1-70b on WikiQA with Verb. prompts. }
    \label{fig:cali_diagram_wikiqa}
\end{figure*}

Similar to Fig. \ref{fig:cali_diagram_plot}, Fig. \ref{fig:cali_diagram_triviaqa}-\ref{fig:cali_diagram_wikiqa} show the reliability diagrams of other three datasets for Llama3.1-70 with Verb. prompts. We find that for high accuracy (>50\%) datasets (TriviaQA and Sciq), Calib-* using BCE and FL loss is less conservative and more likely to predict high confidence. 

\subsection{Detailed Results of LLM Calibration} \label{other_calibration}
Similar to Table \ref{tb:nq_large}, Table \ref{tb:trivial_small}
-\ref{tb:nq_small} show the rest of the detailed calibration results of different methods.

\begin{table*}[ht]
\centering
\scriptsize
\renewcommand{\arraystretch}{.93} 

\begin{tabular}{l@{\hspace{0.1cm}} l@{\hspace{0.15cm}}|l@{\hspace{0.2cm}}l@{\hspace{0.1cm}}l@{\hspace{0.1cm}}l@{\hspace{0.1cm}}|l@{\hspace{0.2cm}}l@{\hspace{0.1cm}}l@{\hspace{0.1cm}}l@{\hspace{0.1cm}}|l@{\hspace{0.2cm}}l@{\hspace{0.1cm}}l@{\hspace{0.1cm}}l@{\hspace{0.1cm}}|l@{\hspace{0.2cm}}l@{\hspace{0.1cm}}l@{\hspace{0.1cm}}l}
\toprule
 & \multirow{3}{*}{\textbf{Method}} & \multicolumn{4}{c}{\textbf{Verb.}} & \multicolumn{4}{c}{\textbf{Zero-shot}} & \multicolumn{4}{c}{\textbf{CoT}} & \multicolumn{4}{c}{\textbf{Few-shot}} \\ 

 \cline{3-18}
 &  & \textbf{\tiny ECE$\downarrow$} & \textbf{\tiny ECE-t$\downarrow$} & \textbf{\tiny Brier $\downarrow$} & \textbf{\tiny AUC$\uparrow$} 
    & \textbf{\tiny ECE $\downarrow$} & \textbf{\tiny ECE-t$\downarrow$} & \textbf{\tiny Brier $\downarrow$} & \textbf{\tiny AUC$\uparrow$} 
    & \textbf{\tiny ECE $\downarrow$} & \textbf{\tiny ECE-t$\downarrow$} & \textbf{\tiny Brier $\downarrow$} & \textbf{\tiny AUC$\uparrow$}  
    & \textbf{\tiny ECE $\downarrow$} & \textbf{\tiny ECE-t$\downarrow$} & \textbf{\tiny Brier $\downarrow$} & \textbf{\tiny AUC$\uparrow$} \\ 
    
\hline
 \multirow{12}{*}{\rotatebox{90}{Llama2-7b}} & LLM Prob. & \cellcolor[HTML]{FEF6BB}0.437 & \cellcolor[HTML]{FEF0B3}0.239 & \cellcolor[HTML]{FEF6BB}0.401 & \cellcolor[HTML]{EC7014}0.737 & \cellcolor[HTML]{FEF6BB}0.296 & \cellcolor[HTML]{F7C37C}0.170 & \cellcolor[HTML]{FDEAAC}0.291 & \cellcolor[HTML]{EC7014}0.746 & \cellcolor[HTML]{FEF6BB}0.314 & \cellcolor[HTML]{F8CB86}0.180 & \cellcolor[HTML]{FEF6BB}0.324 & \cellcolor[HTML]{EE822B}0.693 & \cellcolor[HTML]{FEF6BB}0.419 & \cellcolor[HTML]{EC7014}0.042 & \cellcolor[HTML]{FEF6BB}0.400 & \cellcolor[HTML]{FFF7BC}0.643 \\
\hline \hline 
\end{tabular}
\caption{Test results of small-size models on TriviaQA dataset.}\label{tb:trivial_small}
\end{table*}
\begin{table*}[!ht]
\centering
\scriptsize
\renewcommand{\arraystretch}{.95} 

\begin{tabular}{l@{\hspace{0.1cm}} l@{\hspace{0.15cm}}|l@{\hspace{0.2cm}}l@{\hspace{0.1cm}}l@{\hspace{0.1cm}}l@{\hspace{0.1cm}}|l@{\hspace{0.2cm}}l@{\hspace{0.1cm}}l@{\hspace{0.1cm}}l@{\hspace{0.1cm}}|l@{\hspace{0.2cm}}l@{\hspace{0.1cm}}l@{\hspace{0.1cm}}l@{\hspace{0.1cm}}|l@{\hspace{0.2cm}}l@{\hspace{0.1cm}}l@{\hspace{0.1cm}}l}
\toprule
 & \multirow{3}{*}{\textbf{Method}} & \multicolumn{4}{c}{\textbf{Verb.}} & \multicolumn{4}{c}{\textbf{Zero-shot}} & \multicolumn{4}{c}{\textbf{CoT}} & \multicolumn{4}{c}{\textbf{Few-shot}} \\ 

 \cline{3-18}
 &  & \textbf{\tiny ECE$\downarrow$} & \textbf{\tiny ECE-t$\downarrow$} & \textbf{\tiny Brier $\downarrow$} & \textbf{\tiny AUC$\uparrow$} 
    & \textbf{\tiny ECE $\downarrow$} & \textbf{\tiny ECE-t$\downarrow$} & \textbf{\tiny Brier $\downarrow$} & \textbf{\tiny AUC$\uparrow$} 
    & \textbf{\tiny ECE $\downarrow$} & \textbf{\tiny ECE-t$\downarrow$} & \textbf{\tiny Brier $\downarrow$} & \textbf{\tiny AUC$\uparrow$}  
    & \textbf{\tiny ECE $\downarrow$} & \textbf{\tiny ECE-t$\downarrow$} & \textbf{\tiny Brier $\downarrow$} & \textbf{\tiny AUC$\uparrow$} \\ 
    
\hline

 \multirow{12}{*}{\rotatebox{90}{Gemma2-27b}} & LLM Prob. & \cellcolor[HTML]{FBE1A0}0.294 & \cellcolor[HTML]{FEF6BB}0.251 & \cellcolor[HTML]{F8C780}0.308 & \cellcolor[HTML]{EC7014}0.700 & \cellcolor[HTML]{FEF6BB}0.315 & \cellcolor[HTML]{FEF6BB}0.227 & \cellcolor[HTML]{FCE7A8}0.312 & \cellcolor[HTML]{EC7014}0.729 & \cellcolor[HTML]{FEF4B9}0.257 & \cellcolor[HTML]{FEF6BB}0.254 & \cellcolor[HTML]{FEF1B5}0.307 & \cellcolor[HTML]{FCE3A3}0.593 & \cellcolor[HTML]{F7C179}0.136 & \cellcolor[HTML]{FBDD9B}0.173 & \cellcolor[HTML]{F8C57E}0.226 & \cellcolor[HTML]{FFF7BC}0.563 \\
\hline \hline

\end{tabular}
\caption{Test results of large-size models on TriviaQA dataset.}\label{tb:triviaqa_large}
\end{table*}
\begin{table*}[ht]
\centering
\scriptsize
\renewcommand{\arraystretch}{.93} 

\begin{tabular}{l@{\hspace{0.1cm}} l@{\hspace{0.15cm}}|l@{\hspace{0.2cm}}l@{\hspace{0.1cm}}l@{\hspace{0.1cm}}l@{\hspace{0.1cm}}|l@{\hspace{0.2cm}}l@{\hspace{0.1cm}}l@{\hspace{0.1cm}}l@{\hspace{0.1cm}}|l@{\hspace{0.2cm}}l@{\hspace{0.1cm}}l@{\hspace{0.1cm}}l@{\hspace{0.1cm}}|l@{\hspace{0.2cm}}l@{\hspace{0.1cm}}l@{\hspace{0.1cm}}l}
\toprule
 & \multirow{3}{*}{\textbf{Method}} & \multicolumn{4}{c}{\textbf{Verb.}} & \multicolumn{4}{c}{\textbf{Zero-shot}} & \multicolumn{4}{c}{\textbf{CoT}} & \multicolumn{4}{c}{\textbf{Few-shot}} \\ 

 \cline{3-18}
 &  & \textbf{\tiny ECE$\downarrow$} & \textbf{\tiny ECE-t$\downarrow$} & \textbf{\tiny Brier $\downarrow$} & \textbf{\tiny AUC$\uparrow$} 
    & \textbf{\tiny ECE $\downarrow$} & \textbf{\tiny ECE-t$\downarrow$} & \textbf{\tiny Brier $\downarrow$} & \textbf{\tiny AUC$\uparrow$} 
    & \textbf{\tiny ECE $\downarrow$} & \textbf{\tiny ECE-t$\downarrow$} & \textbf{\tiny Brier $\downarrow$} & \textbf{\tiny AUC$\uparrow$}  
    & \textbf{\tiny ECE $\downarrow$} & \textbf{\tiny ECE-t$\downarrow$} & \textbf{\tiny Brier $\downarrow$} & \textbf{\tiny AUC$\uparrow$} \\ 
    
\hline

 \multirow{12}{*}{\rotatebox{90}{Llama2-7b}} & LLM Prob. & \cellcolor[HTML]{FCE4A4}0.323 & \cellcolor[HTML]{F7C37C}0.196 & \cellcolor[HTML]{F8CA84}0.329 & \cellcolor[HTML]{EC7014}0.712 & \cellcolor[HTML]{FAD490}0.238 & \cellcolor[HTML]{F3A353}0.152 & \cellcolor[HTML]{F4AF63}0.270 & \cellcolor[HTML]{EC7116}0.735 & \cellcolor[HTML]{FEF6BB}0.240 & \cellcolor[HTML]{F4AE62}0.147 & \cellcolor[HTML]{FEF6BB}0.286 & \cellcolor[HTML]{FDE9AA}0.661 & \cellcolor[HTML]{FEF6BB}0.398 & \cellcolor[HTML]{EE8028}0.045 & \cellcolor[HTML]{FEF6BB}0.394 & \cellcolor[HTML]{FFF7BC}0.604 \\
\hline \hline

\end{tabular}
\caption{Test results of small-size models on Sciq dataset.}\label{tb:sciq_small}
\end{table*}
\begin{table*}[ht]
\centering
\scriptsize
\renewcommand{\arraystretch}{.95} 

\begin{tabular}{l@{\hspace{0.1cm}} l@{\hspace{0.15cm}}|l@{\hspace{0.2cm}}l@{\hspace{0.1cm}}l@{\hspace{0.1cm}}l@{\hspace{0.1cm}}|l@{\hspace{0.2cm}}l@{\hspace{0.1cm}}l@{\hspace{0.1cm}}l@{\hspace{0.1cm}}|l@{\hspace{0.2cm}}l@{\hspace{0.1cm}}l@{\hspace{0.1cm}}l@{\hspace{0.1cm}}|l@{\hspace{0.2cm}}l@{\hspace{0.1cm}}l@{\hspace{0.1cm}}l}
\toprule
 & \multirow{3}{*}{\textbf{Method}} & \multicolumn{4}{c}{\textbf{Verb.}} & \multicolumn{4}{c}{\textbf{Zero-shot}} & \multicolumn{4}{c}{\textbf{CoT}} & \multicolumn{4}{c}{\textbf{Few-shot}} \\ 

 \cline{3-18}
 &  & \textbf{\tiny ECE$\downarrow$} & \textbf{\tiny ECE-t$\downarrow$} & \textbf{\tiny Brier $\downarrow$} & \textbf{\tiny AUC$\uparrow$} 
    & \textbf{\tiny ECE $\downarrow$} & \textbf{\tiny ECE-t$\downarrow$} & \textbf{\tiny Brier $\downarrow$} & \textbf{\tiny AUC$\uparrow$} 
    & \textbf{\tiny ECE $\downarrow$} & \textbf{\tiny ECE-t$\downarrow$} & \textbf{\tiny Brier $\downarrow$} & \textbf{\tiny AUC$\uparrow$}  
    & \textbf{\tiny ECE $\downarrow$} & \textbf{\tiny ECE-t$\downarrow$} & \textbf{\tiny Brier $\downarrow$} & \textbf{\tiny AUC$\uparrow$} \\ 
    
\hline

 \multirow{12}{*}{\rotatebox{90}{Gemma2-27b}} & LLM Prob. & \cellcolor[HTML]{FCE2A2}0.193 & \cellcolor[HTML]{FEF1B4}0.181 & \cellcolor[HTML]{F3A455}0.236 & \cellcolor[HTML]{EC7014}0.705 & \cellcolor[HTML]{FCE5A6}0.204 & \cellcolor[HTML]{FCE3A3}0.179 & \cellcolor[HTML]{F29C4B}0.229 & \cellcolor[HTML]{EC7014}0.713 & \cellcolor[HTML]{F9CE8A}0.161 & \cellcolor[HTML]{F9D08C}0.158 & \cellcolor[HTML]{F7BE75}0.227 & \cellcolor[HTML]{F7C27A}0.630 & \cellcolor[HTML]{F29B4A}0.080 & \cellcolor[HTML]{F5B367}0.088 & \cellcolor[HTML]{F5B063}0.172 & \cellcolor[HTML]{FFF7BC}0.616 \\
\hline \hline

\end{tabular}
\caption{Test results of large-size models on Sciq dataset.}
\end{table*}
\begin{table*}[ht]
\centering
\scriptsize
\renewcommand{\arraystretch}{.93} 

\begin{tabular}{l@{\hspace{0.1cm}} l@{\hspace{0.15cm}}|l@{\hspace{0.2cm}}l@{\hspace{0.1cm}}l@{\hspace{0.1cm}}l@{\hspace{0.1cm}}|l@{\hspace{0.2cm}}l@{\hspace{0.1cm}}l@{\hspace{0.1cm}}l@{\hspace{0.1cm}}|l@{\hspace{0.2cm}}l@{\hspace{0.1cm}}l@{\hspace{0.1cm}}l@{\hspace{0.1cm}}|l@{\hspace{0.2cm}}l@{\hspace{0.1cm}}l@{\hspace{0.1cm}}l}
\toprule
 & \multirow{3}{*}{\textbf{Method}} & \multicolumn{4}{c}{\textbf{Verb.}} & \multicolumn{4}{c}{\textbf{Zero-shot}} & \multicolumn{4}{c}{\textbf{CoT}} & \multicolumn{4}{c}{\textbf{Few-shot}} \\ 

 \cline{3-18}
 &  & \textbf{\tiny ECE$\downarrow$} & \textbf{\tiny ECE-t$\downarrow$} & \textbf{\tiny Brier $\downarrow$} & \textbf{\tiny AUC$\uparrow$} 
    & \textbf{\tiny ECE $\downarrow$} & \textbf{\tiny ECE-t$\downarrow$} & \textbf{\tiny Brier $\downarrow$} & \textbf{\tiny AUC$\uparrow$} 
    & \textbf{\tiny ECE $\downarrow$} & \textbf{\tiny ECE-t$\downarrow$} & \textbf{\tiny Brier $\downarrow$} & \textbf{\tiny AUC$\uparrow$}  
    & \textbf{\tiny ECE $\downarrow$} & \textbf{\tiny ECE-t$\downarrow$} & \textbf{\tiny Brier $\downarrow$} & \textbf{\tiny AUC$\uparrow$} \\ 
    
\hline

 \multirow{12}{*}{\rotatebox{90}{Llama2-7b}} & LLM Prob. & \cellcolor[HTML]{FEF4B8}0.426 & \cellcolor[HTML]{FBE1A1}0.207 & \cellcolor[HTML]{FEF2B6}0.392 & \cellcolor[HTML]{EE842D}0.692 & \cellcolor[HTML]{FEF6BB}0.369 & \cellcolor[HTML]{F8C780}0.140 & \cellcolor[HTML]{FEF6BB}0.301 & \cellcolor[HTML]{EC7115}0.751 & \cellcolor[HTML]{FEF6BB}0.248 & \cellcolor[HTML]{F9D18C}0.153 & \cellcolor[HTML]{FEF6BB}0.291 & \cellcolor[HTML]{FFF7BC}0.620 & \cellcolor[HTML]{FEF6BB}0.409 & \cellcolor[HTML]{F4AC5E}0.117 & \cellcolor[HTML]{FEF6BB}0.402 & \cellcolor[HTML]{FFF7BC}0.578 \\
\hline \hline

\end{tabular}
\caption{Test results of small-size models on WikiQA dataset.}
\end{table*}
\begin{table*}[ht]
\centering
\scriptsize
\renewcommand{\arraystretch}{.95} 

\begin{tabular}{l@{\hspace{0.1cm}} l@{\hspace{0.15cm}}|l@{\hspace{0.2cm}}l@{\hspace{0.1cm}}l@{\hspace{0.1cm}}l@{\hspace{0.1cm}}|l@{\hspace{0.2cm}}l@{\hspace{0.1cm}}l@{\hspace{0.1cm}}l@{\hspace{0.1cm}}|l@{\hspace{0.2cm}}l@{\hspace{0.1cm}}l@{\hspace{0.1cm}}l@{\hspace{0.1cm}}|l@{\hspace{0.2cm}}l@{\hspace{0.1cm}}l@{\hspace{0.1cm}}l}
\toprule
 & \multirow{3}{*}{\textbf{Method}} & \multicolumn{4}{c}{\textbf{Verb.}} & \multicolumn{4}{c}{\textbf{Zero-shot}} & \multicolumn{4}{c}{\textbf{CoT}} & \multicolumn{4}{c}{\textbf{Few-shot}} \\ 

 \cline{3-18}
 &  & \textbf{\tiny ECE$\downarrow$} & \textbf{\tiny ECE-t$\downarrow$} & \textbf{\tiny Brier $\downarrow$} & \textbf{\tiny AUC$\uparrow$} 
    & \textbf{\tiny ECE $\downarrow$} & \textbf{\tiny ECE-t$\downarrow$} & \textbf{\tiny Brier $\downarrow$} & \textbf{\tiny AUC$\uparrow$} 
    & \textbf{\tiny ECE $\downarrow$} & \textbf{\tiny ECE-t$\downarrow$} & \textbf{\tiny Brier $\downarrow$} & \textbf{\tiny AUC$\uparrow$}  
    & \textbf{\tiny ECE $\downarrow$} & \textbf{\tiny ECE-t$\downarrow$} & \textbf{\tiny Brier $\downarrow$} & \textbf{\tiny AUC$\uparrow$} \\ 
    
\hline
 \multirow{12}{*}{\rotatebox{90}{Gemma2-27b}} & LLM Prob. & \cellcolor[HTML]{F9CF8B}0.423 & \cellcolor[HTML]{FEF6BB}0.258 & \cellcolor[HTML]{F7C077}0.382 & \cellcolor[HTML]{EC7014}0.775 & \cellcolor[HTML]{FEF6BB}0.448 & \cellcolor[HTML]{FEF6BB}0.259 & \cellcolor[HTML]{FEF6BB}0.410 & \cellcolor[HTML]{F9CF8A}0.730 & \cellcolor[HTML]{FEF6BB}0.359 & \cellcolor[HTML]{FEF6BB}0.222 & \cellcolor[HTML]{FEF6BB}0.356 & \cellcolor[HTML]{FBDF9E}0.654 & \cellcolor[HTML]{FEF6BB}0.158 & \cellcolor[HTML]{FEF6BB}0.164 & \cellcolor[HTML]{FEF6BB}0.226 & \cellcolor[HTML]{FFF7BC}0.501 \\
\hline \hline

\end{tabular}
\caption{Test results of large-size models on WikiQA dataset.}
\end{table*}
\begin{table*}[ht]
\centering
\scriptsize
\renewcommand{\arraystretch}{.93} 

\begin{tabular}{l@{\hspace{0.1cm}} l@{\hspace{0.15cm}}|l@{\hspace{0.2cm}}l@{\hspace{0.1cm}}l@{\hspace{0.1cm}}l@{\hspace{0.1cm}}|l@{\hspace{0.2cm}}l@{\hspace{0.1cm}}l@{\hspace{0.1cm}}l@{\hspace{0.1cm}}|l@{\hspace{0.2cm}}l@{\hspace{0.1cm}}l@{\hspace{0.1cm}}l@{\hspace{0.1cm}}|l@{\hspace{0.2cm}}l@{\hspace{0.1cm}}l@{\hspace{0.1cm}}l}
\toprule
 & \multirow{3}{*}{\textbf{Method}} & \multicolumn{4}{c}{\textbf{Verb.}} & \multicolumn{4}{c}{\textbf{Zero-shot}} & \multicolumn{4}{c}{\textbf{CoT}} & \multicolumn{4}{c}{\textbf{Few-shot}} \\ 

 \cline{3-18}
 &  & \textbf{\tiny ECE$\downarrow$} & \textbf{\tiny ECE-t$\downarrow$} & \textbf{\tiny Brier $\downarrow$} & \textbf{\tiny AUC$\uparrow$} 
    & \textbf{\tiny ECE $\downarrow$} & \textbf{\tiny ECE-t$\downarrow$} & \textbf{\tiny Brier $\downarrow$} & \textbf{\tiny AUC$\uparrow$} 
    & \textbf{\tiny ECE $\downarrow$} & \textbf{\tiny ECE-t$\downarrow$} & \textbf{\tiny Brier $\downarrow$} & \textbf{\tiny AUC$\uparrow$}  
    & \textbf{\tiny ECE $\downarrow$} & \textbf{\tiny ECE-t$\downarrow$} & \textbf{\tiny Brier $\downarrow$} & \textbf{\tiny AUC$\uparrow$} \\ 
    
\hline
 \multirow{12}{*}{\rotatebox{90}{Llama2-7b}} & LLM Prob. & \cellcolor[HTML]{FEF6BB}0.554 & \cellcolor[HTML]{FEF6BB}0.248 & \cellcolor[HTML]{FEF6BB}0.480 & \cellcolor[HTML]{EC7014}0.700 & \cellcolor[HTML]{FEF6BB}0.436 & \cellcolor[HTML]{FADA98}0.182 & \cellcolor[HTML]{FEF6BB}0.356 & \cellcolor[HTML]{EC7318}0.719 & \cellcolor[HTML]{FEF6BB}0.488 & \cellcolor[HTML]{FAD592}0.182 & \cellcolor[HTML]{FEF6BB}0.445 & \cellcolor[HTML]{FDEBAE}0.647 & \cellcolor[HTML]{FEF6BB}0.543 & \cellcolor[HTML]{F08F3A}0.047 & \cellcolor[HTML]{FEF6BB}0.477 & \cellcolor[HTML]{FFF7BC}0.628 \\
\hline \hline

\end{tabular}
\caption{Test results of small-size models on NQ dataset.}\label{tb:nq_small}
\end{table*}

\end{document}